\documentclass[lettersize,journal]{IEEEtran}
\usepackage{amsmath,amsfonts}
\usepackage{algorithmic}
\usepackage{array}
\usepackage{textcomp}
\usepackage{stfloats}
\usepackage{url}
\usepackage{verbatim}
\usepackage{graphicx}
\usepackage{pifont}
\newcommand{\cmark}{\ding{51}}%
\newcommand{\xmark}{\ding{55}}%
\usepackage{multirow}
\usepackage{booktabs}
\usepackage{arydshln}
\usepackage{comment}
\usepackage{enumitem}
\usepackage{subfigure}
\usepackage{algorithm}
\usepackage{color}
\usepackage{listings}
\definecolor{lightgray}{gray}{0.9}

\lstdefinestyle{mystyle}{
    language=Python,
    basicstyle=\ttfamily\small,
    keywordstyle=\color{black},
    commentstyle=\color{black},
    stringstyle=\color{black},
    breakatwhitespace=false,         
    breaklines=true,                 
    captionpos=b,                    
    keepspaces=true,                 
    showspaces=false,                
    showstringspaces=false,
    showtabs=false,                  
    tabsize=2
}

\def\BibTeX{{\rm B\kern-.05em{\sc i\kern-.025em b}\kern-.08em
    T\kern-.1667em\lower.7ex\hbox{E}\kern-.125emX}}
\usepackage{balance}
\begin{document}
\title{\textsc{\textbf{Q-Bench$^+$}}:\\ A Benchmark for Multi-modal Foundation Models on Low-level Vision from Single Images to Pairs}
\author{Zicheng Zhang*, Haoning Wu*, Erli Zhang,\\ Guangtao Zhai$^{\dagger}$, \emph{Senior Member, IEEE,} and Weisi Lin$^{\dagger}$, \emph{Fellow, IEEE}
\IEEEcompsocitemizethanks{\IEEEcompsocthanksitem This work was supported in part by the College of Engineering (CoE) Research Award 2023, Nanyang Technological University, Grant No: 022877, and the National Natural Science Foundation of China No: (623B2073, 62101326, 62225112, 62301316). \protect}
\IEEEcompsocitemizethanks{\IEEEcompsocthanksitem Zicheng Zhang and Guangtao Zhai are with the Institute of Image Communication and Network Engineering, Shanghai Jiao Tong University, 200240 Shanghai, China. E-mail:\{zzc1998,zhaiguangtao\}
@sjtu.edu.cn. \protect}
\IEEEcompsocitemizethanks{\IEEEcompsocthanksitem Haoning Wu and Erli Zhang are
with S-Lab, Nanyang Technological University, Singapore. E-mail: \{haoning001,ezhang005\}@e.ntu.edu.sg. \protect}
\IEEEcompsocitemizethanks{\IEEEcompsocthanksitem Weisi Lin is with the School of Computer Science and Engineering, Nanyang Technological University, Singapore.  E-mail: wslin@ntu.edu.sg.\protect}
\IEEEcompsocitemizethanks{\IEEEcompsocthanksitem 
*Equal Contributions. $^{\dagger}$Corresponding Authors. \protect}}


\maketitle

\begin{abstract}
The rapid development of Multi-modality Large Language Models (MLLMs) has navigated a paradigm shift in computer vision, moving towards versatile foundational models. However, evaluating MLLMs in \textit{low-level visual perception and understanding} remains a yet-to-explore domain. To this end, we design benchmark settings to \textit{emulate human language responses} related to low-level vision: the low-level visual \emph{perception} (\underline{A1}) \textit{via} visual question answering related to low-level attributes (\textit{e.g.~clarity, lighting}); and the low-level visual \emph{description} (\underline{A2}), on evaluating MLLMs for low-level text descriptions. Furthermore, given that pairwise comparison can better avoid ambiguity of responses and has been adopted by many human experiments, we further extend the low-level perception-related question-answering and description evaluations of MLLMs from single images to \textit{image pairs}. Specifically, for \textit{perception} (A1), we carry out the LLVisionQA$^{+}$ dataset, comprising 2,990 single images and 1,999 image pairs each accompanied by an open-ended question about its low-level features; for \textbf{\textit{description}} (A2), we propose the LLDescribe$^{+}$ dataset, evaluating MLLMs for low-level descriptions on 499 single images and 450 pairs. Additionally, we evaluate MLLMs on \textbf{\textit{assessment}} (A3) ability, \textit{i.e.} predicting score, by employing a softmax-based approach to enable all MLLMs to generate \textit{quantifiable} quality ratings, tested against human opinions in 7 image quality assessment (IQA) datasets. With 24 MLLMs under evaluation, we demonstrate that several MLLMs have decent low-level visual competencies on single images, but only GPT-4V exhibits higher accuracy on pairwise comparisons than single image evaluations (\textit{like humans}). We hope that our benchmark will motivate further research into uncovering and enhancing these nascent capabilities of MLLMs. Datasets will be available at \url{https://github.com/Q-Future/Q-Bench}.
\end{abstract}

\begin{IEEEkeywords}
Multi-modality large language models (MLLM), low-level vision, benchmark, perception, description, assessment
\end{IEEEkeywords}

\section{Introduction}

\begin{figure}
    \centering
    \includegraphics[width=0.8\linewidth]{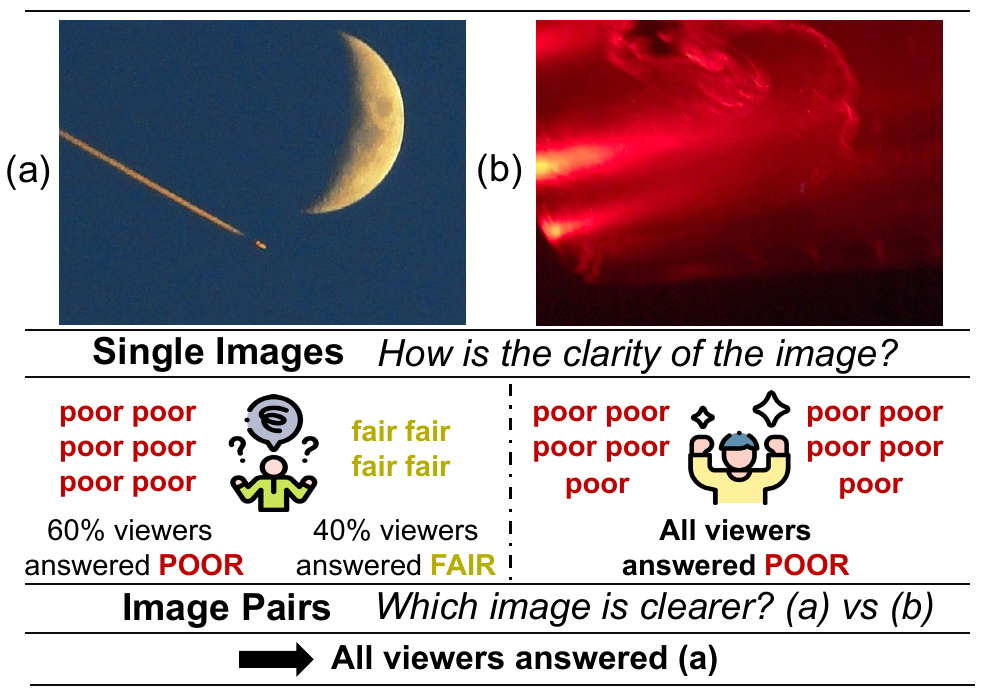}
    \vspace{-1em}
    \caption{\textbf{Pairwise comparison} is a non-negligible setting for human to perceive and evaluate low-level visual attributes, as it provides additional and non-ambiguous information (\textit{\textbf{(a)} is clearer than \textbf{(b)}}). Henceforth, we extend into the \textbf{Q-Bench$^+$} with image pairs to examine whether MLLMs can \textit{extract and compare low-level visual information between a pair of images, like a human}.}
    \label{fig:singlevspair}
    \vspace{-1.5em}
\end{figure}
\IEEEPARstart{T}he emergent large language models (LLMs) such as ChatGPT and Bard, as well as their excellent open-source counterparts (\textit{e.g.}, LLaMA~\cite{llama}, MPT~\cite{mpt}), have served as powerful general-purpose assistants, which opens a new era for artificial intelligence (AI) from targeting specific tasks towards general intelligence. Following the advancements of LLMs, multi-modality large language models (MLLMs), as represented by LLaVA~\cite{llava}, MiniGPT-4~\cite{minigpt4}, InstructBLIP~\cite{iblip}, and Otter~\cite{otter}, have brought exciting progresses on the vision field as well. They are capable of providing robust general-level abilities on visual perception/understanding and can even seamlessly dialog and interact with humans through natural language. While such abilities of MLLMs have been validated on several vision-language tasks such as image captioning~\cite{cococaps}, visual question answering~\cite{cocovqa}, cross-modality grounding~\cite{kosmos2}, and traditional vision tasks such as image classification or segmentation~\cite{lai2023lisa}, most attention is paid to the high-level perception and understanding of visual contents. Meanwhile, the ability of MLLMs remains unclear on \textbf{low-level visual perception and understanding}, which play significant roles in image quality assessment (IQA)~\cite{koniq,spaq} and its associated tasks on perceiving visual distortions (\textit{noises, blurs})~\cite{koniqplusplus,wu2023explainable}, and other low-level attributes (\textit{color, lighting, composition, style, etc})~\cite{aadb} that may relate to aesthetics of natural photos~\cite{avaiaa} as well as human preferences on emerging computer-graphics generated~\cite{zhang2023subjective} or AI-generated images~\cite{agiqa3k,imagereward}. These low-level visual abilities are strongly associated with a wide range of applications, such as recommendation~\cite{wu2023dover}, guidance on camera systems~\cite{irpotential}, or visual quality enhancement~\cite{lpips}. Henceforth, it is crucial to evaluate these general-purpose foundation models in low-level visual perception and understanding, to relieve extensive human resources on giving feedback to every specific low-level task.

\begin{figure*}
    \centering
    \includegraphics[width=.95\linewidth]{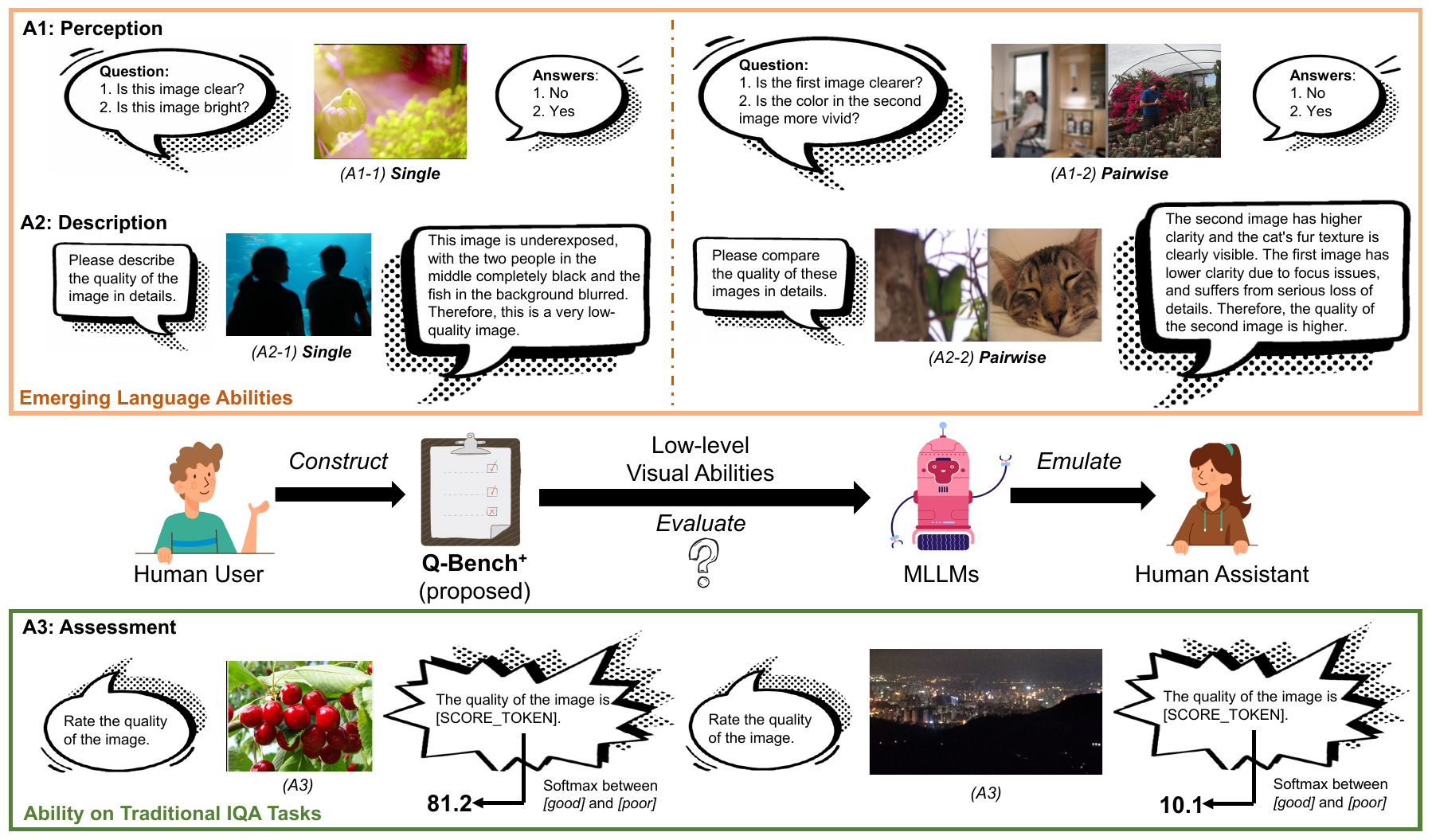}
    \vspace{-10pt}
    \caption{In the proposed \textbf{Q-Bench$^+$}, we build the first benchmark on emerging abilities of MLLMs for low-level vision, including \textbf{perception} of single/pairwise low-level attributes (\textit{by correctly answering diverse queries}) and \textbf{description} of single/pairwise low-level quality-related information via natural language. Furthermore, the \textbf{Q-Bench$^+$} also evaluates the \textit{quantifiable} \textbf{assessment} ability of MLLMs on traditional IQA tasks.}
    \vspace{-10pt}
    \label{fig:1}

\end{figure*}

In this paper, we propose the first systematic benchmark \textbf{Q-Bench$^+$} to measure the low-level visual abilities of MLLMs, which is constructed around a key question:

\textit{How do MLLMs emulate human ability related to low-level visual perception and understanding?}

A basic answer comes from the fundamental capability of MLLMs: \textbf{vision-conditioned language generation}. Specifically, for low-level vision, MLLMs should ideally be able to correctly answer low-level visual questions and precisely describe the low-level information of single images. Henceforth, we define the following two emerging abilities of MLLMs that directly arise from their language generation capability:

\textit{\textbf{Ability 1: Perception} of Low-level Attributes.} As shown in Fig.~\ref{fig:1} (A1-1), like a human, an MLLM should be able to respond accurately to simple questions related to low-level attributes, \textit{e.g} answering \textit{`No'} for a blurry image when queried with \textit{`Is this image clear?'}.

 \textit{\textbf{Ability 2: Description} via Natural Language}. As shown in Fig.~\ref{fig:1} (A2-1), like a human, an MLLM should be able to describe the quality and other low-level related attributes for single images with natural language. The descriptions should be both complete and accurate.

Although the above two capabilities essentially emulate human perception of low-level vision, they still miss some key capabilities of humans. For example, regarding Fig.~\ref{fig:singlevspair} (a), some people may consider its clarity to be average, while others may deem it poor, while neither opinion should be considered incorrect; instead, everyone would agree that Fig.~\ref{fig:singlevspair} (b) \textit{is clearer than} Fig.~\ref{fig:singlevspair} (a). On the other hand, for those who regard both Fig.~\ref{fig:singlevspair} (b) and Fig.~\ref{fig:singlevspair} (a) as blurry, comparing the clarity between the pair can also provide additional valuable information. Noticing these issues, lots of recent subjective studies~\cite{lpips,pieapp,pipal} have adopted the a juxtaposition-based paradigm, that is, collecting human opinions by comparing \textbf{a pair of images}. Based on these insights and recent progresses on MLLMs~\cite{emu2,bakllava,geminipro,openai2023gpt4} that officially support more than one images as inputs, we further explore whether MLLMs can similarly emulate respective human capabilities:

\textit{Can MLLMs adeptly extract and compare low-level visual information between a pair of images?}

On answering this question, we further extend the \textbf{Perception} and \textbf{Description} tasks from single images to image pairs:

\textit{\textbf{Extended Ability 1: Perception} of Low-level Attributes for \textbf{image pairs}.} As shown in Fig.~\ref{fig:1} (A1-2), like a human, an MLLM should be able to respond correctly to low-level questions for image pairs, \textit{e.g} answering \textit{`No'} for image pair (first blurrier) when queried with \textit{`Is the first image clearer?'}. 

\textit{\textbf{Extended Ability 2: Description} via Natural Language for \textbf{image pairs}}. As shown in Fig.~\ref{fig:1} (A2-2), like a human, an MLLM should be able to describe the similarities (\textit{joint information}) and differences (\textit{comparison}) of low-level appearances between a pair of images with natural language.

Despite the direct and concrete abilities above, we also evaluate how MLLMs can perform on the traditional IQA task, a highly abstract task that requires understanding on how the low-level attributes  affect human judgements, as follows:

\textit{\textbf{Ability 3:} Precise \textbf{Assessment} Aligned with Human Opinions.} As depicted in Fig.~\ref{fig:1} (A3), an MLLM should be able to predict \textit{quantifiable} quality scores for images, which can be aligned with the human-rated mean opinion scores (MOS).

To evaluate the three abovementioned abilities, we formulate their respective benchmark settings, as follows:

\subsubsection{\textbf{LLVisionQA$^+$ Benchmark Dataset}}
To evaluate the low-level {\textbf{perception}} ability (A1) on various low-level attributes under diverse circumstances, we construct the \textbf{LLVisionQA$^+$} dataset, including 2,990 single images and 1,999 image pairs from 10 diverse sources. Aligned with existing practices~\cite{mmbench,emabench}, each single image or image pair in \textbf{LLVisionQA$^+$} is equipped with a question, alongside a correct answer and false candidate answers. Specifically, we design three diverse types of questions: \textit{Yes-or-No} questions, \textit{What} questions, and \textit{How} questions. Moreover, we divide low-level concerns for single images into four quadrants, via two axes: (\textbf{1)} distortions (\textit{blur, noises, etc}) \textit{vs} other low-level attributes (\textit{color, lighting, composition, etc})~\cite{atqa}. \textbf{(2)} global perception (\textit{e.g., sharpness of the whole picture}) \textit{vs} local content-related in-context perception (\textit{e.g., whether the red flower is in focus})~\cite{sfa}. On the other hand, we separate the low-level concerns for image pairs into four sub-categories: (\textbf{1)} distortions \textit{vs} other low-level attributes (\textit{similar as above}). \textbf{(2)} comparison (\textit{e.g., which image is clearer}) \textit{vs} joint analysis (\textit{e.g., are both images underexposure}). With three types of questions and divided concerns, the proposed \textbf{LLVisionQA$^+$} dataset provides a holistic benchmark for the low-level perception abilities of MLLMs on both single images and pairs.

\subsubsection{\textbf{LLDescribe$^+$ Benchmark Dataset}}
For the \textbf{description} ability (A2), given that the output description is expected to be open-ended, we propose the \textbf{LLDescribe$^+$} dataset by inviting experts with rich experience in the low-level vision field to write long \textit{golden} low-level descriptions (\textit{average \textbf{58} words per description}) for 499 single images and 450 image pairs. The long \textit{golden} low-level descriptions then serve as the reference texts for the single-modal GPT to evaluate MLLM output descriptions. To ensure the evaluation is comprehensive, the quality of MLLM descriptions is evaluated through three dimensions: completeness (\textit{punish missing information}), preciseness (\textit{punish outputs controversial with reference}), as well as relevance (\textit{punish outputs irrelevant to low-level attributes}). 
With \textit{golden} descriptions and the multi-dimensional evaluation process participated by GPT, we comprehensively evaluate the low-level description ability of MLLMs.

\subsubsection{\textbf{IQA Benchmark}}
For the \textbf{assessment} ability, we utilize plenty of existing IQA databases~\cite{koniq,kadid,agiqa3k,zhang2023subjective,spaq,livechallenge} that focus on various low-level appearances of images, to benchmark MLLMs within conventional IQA settings.
Specifically, we notice that MLLMs encounter difficulties in providing sufficiently \textit{quantifiable} outputs, whether instructed to directly rate with texts or provide numerical outputs. To solve this challenge, we propose to extract the {\tt softmax} pooling result on the logits of the two most frequent tokens (\textbf{\textit{good}} and \textbf{\textit{poor}}) under the response template of MLLMs (Fig.~\ref{fig:1} (A3)) as their quality predictions. {Our studies prove that the proposed {softmax-based} strategy is generally better correlated with human perception than direct token outputs of MLLMs (via {\tt argmax}), which bridges between these emergent MLLMs and the traditional IQA task settings.} Under this strategy, we evaluate all MLLMs on their precise {assessment} ability by measuring the correlations between their predictions and human opinion scores in various IQA databases. Furthermore, we propose a \textbf{prompt-ensemble} approach to help boost the IQA performance of MLLMs with the {softmax-based} strategy.

This work is a substantially extended version of our earlier conference publication~\cite{wu2024qbench}. Compared with the conference version, we bring three major changes: \textbf{(1)} Most importantly, we extend the \textbf{perception} and \textbf{description} tasks from single images to image pairs, which provides a more comprehensive benchmark for MLLMs on emulating human low-level visual understanding ability. \textbf{(2)} We update the benchmark with the latest popular MLLMs (evaluated MLLMs increased from 15 to \textbf{24}), providing a review of the development for MLLMs on low-level vision. \textbf{(3)} We further propose a simple yet effective prompt-ensemble approach, which can help boost the zero-shot performance of MLLMs on the \textbf{assessment} task.

{In summary, we systematically explore the potential of MLLMs on three low-level visual abilities: \textbf{perception}, \textbf{description}, and \textbf{assessment}. The three realms compose into the proposed \textbf{Q-Bench$^+$}, a MLLM benchmark on low-level visual tasks. Our contributions can be summarized as three-fold:}

\begin{itemize}[itemsep=2pt,topsep=0pt,parsep=0pt]
    \item We build a benchmark for MLLMs on low-level \textbf{perception} ability. To achieve this, we construct a first-of-its-kind balanced and comprehensive \textbf{LLVisionQA$^+$} dataset with 2,990 single images and 1,999 image pairs with one low-level-related question-answer pair for each image. The \textbf{LLVisionQA$^+$} dataset includes three question types and multiple low-level concerns to ensure diversity.
    \item We define a benchmark process to evaluate the low-level \textbf{description} ability of MLLMs, including an \textbf{LLDescription$^+$} dataset of 499 single images and 450 image pairs with expert-labeled long \textit{golden} quality descriptions, and a GPT-assisted evaluation to rate MLLM-descriptions in terms of completeness, preciseness, and relevance compared with \textit{golden} descriptions.
    \item To evaluate precise quality \textbf{assessment} ability, we propose a unified {softmax-based} quality prediction strategy for all MLLMs based on their probability outputs. Furthermore, we propose a prompt-ensemble approach to help boost the IQA performance of MLLMs with the {softmax-based} strategy. With its effectiveness validated in our experiments, the proposed strategy sets up a bridge between general-purpose MLLMs and traditional IQA tasks that requires \textit{quantifiable} scores as outputs.  
\end{itemize}

\begin{table}[!t]\small
    \centering
    \renewcommand\arraystretch{1.2}
    \renewcommand\tabcolsep{3pt}
    \caption{Overview of the 10 diverse image source datasets in the \textbf{Q-Bench$^+$}, and the respective benchmark dataset size for each low-level ability among \textbf{perception}, \textbf{descrption} and \textbf{assessment}. The \textit{Corrupted} COCO denotes COCO-Captions images corrupted by \cite{imagecorruptions}.}
    \vspace{-6pt}
   \resizebox{\linewidth}{!}{\begin{tabular}{l|l|c|c|c}
    \toprule
    \multirow{2}{35pt}{\textbf{Type}} & \multirow{2}{*}{\textbf{Source Dataset}} & \textbf{LLVisionQA$^+$} & \textbf{LLDescribe$^+$} & {Full Dataset Size} \\
    & & {Sampled Size}  & {Sampled Size}  & for A3 Task \\ \hline
    \multirow{4}{*}{In-the-wild} & KONiQ-10K~\cite{koniq} & 600 & 200 & 10,073 \\
    & SPAQ~\cite{spaq} & 800 & 200 & 11,125 \\
    & LIVE-FB~\cite{paq2piq} & 300 & 50 & 39,810 \\
    & LIVE-itw~\cite{clive} & 300 & 50 & 1,169 \\ \hdashline
    \multirow{3}{35pt}{Generated} & CGIQA-6K~\cite{zhang2023subjective} & 200 & 50 & 6,000 \\
    & AGIQA-3K~\cite{agiqa3k} & 198 & 80 & 2,982 \\
    & ImageRewardDB~\cite{imagereward} & 194 & 29 & \textit{excluded in} (A3) \\ \hdashline
    \multirow{3}{35pt}{Manually-distorted} & KADID-10K~\cite{kadid} & 81 & 20 & 10,125 \\
    & LIVEMultiDistortion~\cite{livemultipledistortions} & 15 & 10 & \textit{excluded in} (A3) \\
    & \textit{Corrupted} COCO~\cite{cococaps} & 302 & 50 & \textit{excluded in} (A3) \\ \hline

    \multicolumn{2}{c|}{{Corresponding Task in} \textbf{Q-Bench$^+$}} & (A1)~\textbf{Perception} & (A2)~\textbf{Description} & (A3)~\textbf{Assessment} \\
    \multicolumn{2}{c|}{{Benchmark Size (single+pairwise)}} & 2,990+1,999 & 499+450 & 81,284 \\
    \bottomrule
\end{tabular}}
\vspace{-12pt}
    \label{tab:1}
\end{table}

\section{Constructing the Q-Bench$^+$}

\subsection{General Principles}
\label{sec:21}

\subsubsection{Focusing on Low-level Visual Abilities of MLLMs} Unlike existing MLLM benchmarks~\cite{seedbench, mmbench, emabench} that aim at all-round abilities, the tasks in \textbf{Q-Bench$^+$} are constrained with two basic principles: a) Requiring perception and/or understanding on low-level attributes of images; b) Not requiring reasoning (\textit{i.e. why}) or {outside} knowledge~\cite{okvqa}. We adhere to the principles in designing the \textbf{perception}, \textbf{description}, and \textbf{assessment} tasks, making the proposed \textbf{Q-Bench$^+$} a focused reflection on the low-level visual abilities of MLLMs.

\subsubsection{Covering Diverse Low-level Appearances} To cover diverse low-level appearances, we collect multi-sourced images for each task, as depicted in Table~\ref{tab:1}. Among all images in the \textbf{perception} and \textbf{description} tasks, {\textit{two-thirds}} are in-the-wild images directly collected from social media posts, smartphones, or professional photography. The rest {\textit{one-third}} images are collected after various artificial distortions, or via generative processes (CGI, AIGC). Furthermore, we employ k-means clustering for the low-level attribute indicators to certify that the sub-sampled images retain high diversity. In the \textbf{assessment} task, full images of 7 IQA datasets within all three source types are evaluated through traditional IQA metrics. The diverse and multiple sources of images morph the \textbf{Q-Bench$^+$} into a holistic and balanced benchmark to fairly evaluate low-level-related abilities.

\subsubsection{Extending from Single Images to Image Pairs} 
Evaluating image pairs allows for direct comparison and joint analysis of low-level attributes, which can highlight subtle differences or similarities that might not be evident when images are viewed in isolation. Humans are good at comparing, therefore we believe it is also important to benchmark the \textit{low-level visual perception and understanding ability} of MLLMs on image pairs. Thus we extend the benchmark (only including single images) in our conference version \cite{wu2024qbench} with image pairs to simulate more complex visual tasks that mirror real-world scenarios and challenge the MLLMs to process and compare multiple visual inputs simultaneously.

\subsection{Benchmark on Low-level \textbf{Perception} Ability} 
\begin{figure*}
    \centering
    \includegraphics[width=0.95\linewidth]{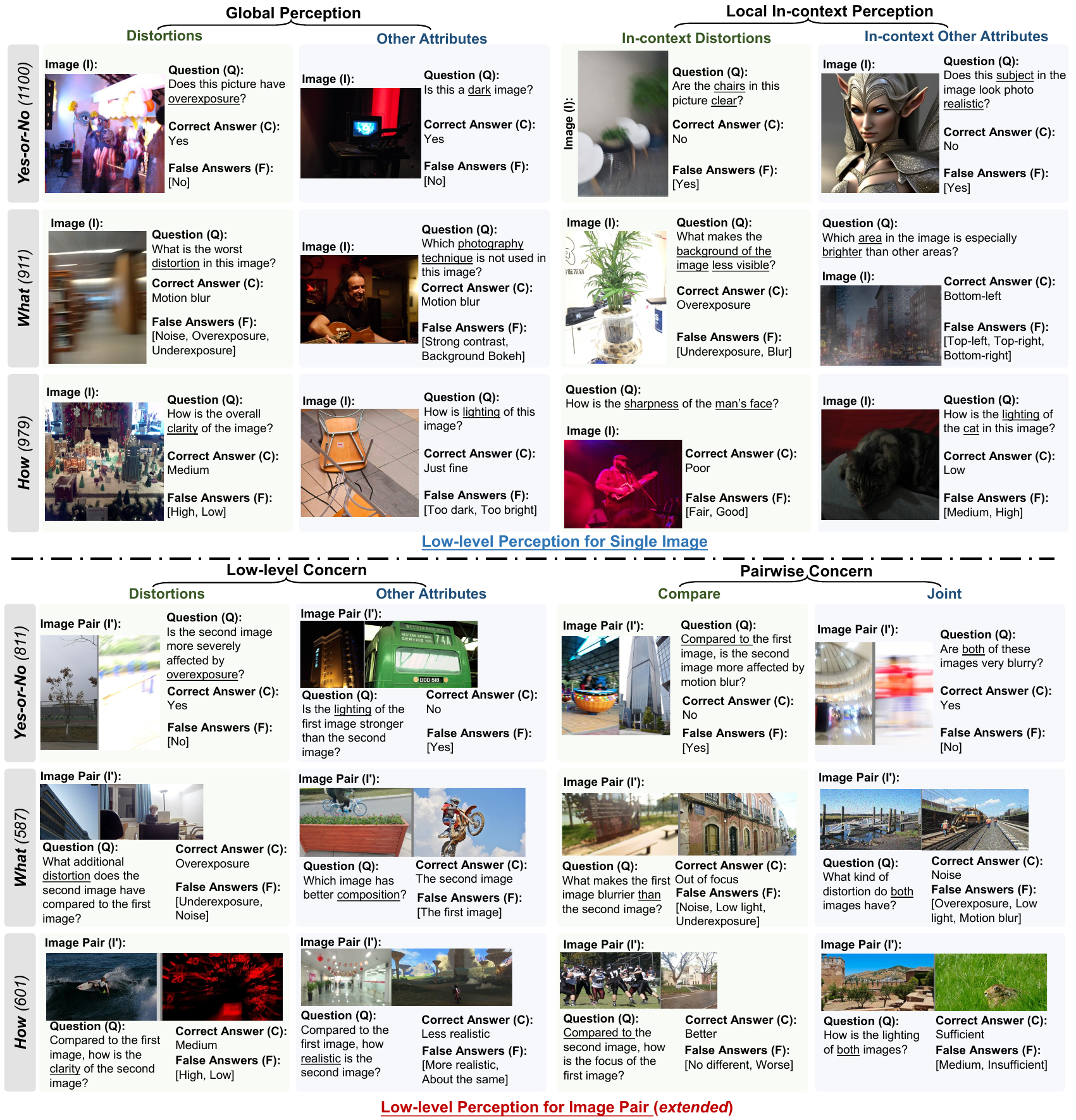}
    \vspace{-8pt}
    \caption{A dataset card of \textbf{LLVisionQA$^+$} that evaluates the low-level \textbf{perception} ability of MLLMs. 2,990 {\tt (I,Q,C,F)} and 1,999 {\tt (I',Q,C,F)} tuples are collected to cover three question types and various low-level visual concerns, providing an all-around evaluation of low-level visual perception for MLLMs.}
    \label{fig:2}
    \vspace{-15pt}
\end{figure*}

In the first task of \textbf{Q-Bench$^+$}, we evaluate the low-level \textbf{perception} ability of MLLMs to examine whether they can answer simple natural queries related to low-level attributes. For this purpose, we first collect 2,990 single images ({\tt I}) from multiple sources (see Table~\ref{tab:1}) with diverse low-level concerns, from which we collect 1,999 image pairs ({\tt I'}) as well. All image pairs are different from each other but may have one repeated image across different pairs. Then, we collect one low-level-related question ({\tt Q}), one correct answer to the question ({\tt C}), and 1-3 candidate false answers ({\tt F}) for each single image or image pair. The 2,990 {\tt (I,Q,C,F)} and 1,999 {\tt (I',Q,C,F)} tuples compose into the \textbf{LLVisionQA$^+$} dataset (as illustrated in Fig.~\ref{fig:2}), the first visual question answering (VQA) dataset in the low-level computer vision field. Specifically, the questions in \textbf{LLVisionQA$^+$} cover four quadrants of distinct low-level concerns and three question types. After constructing the dataset, the {\tt (I,Q,C,F)} are together fed into MLLMs for evaluation, while their outputs are further examined by GPT to judge correctness. The details are elaborated as follows.

\textit{1) Low-level Visual Concerns for Single Images}
\label{sec:221}

\textbf{Axis 1: Distortions \textit{vs} Other Low-level Attributes.} 
The primary axis differentiates two categories of low-level perceptual attributes: \textbf{1)} technical \textbf{distortions}~\cite{koniqplusplus}, seen as the low-level characteristics that directly degrade the quality of images~\cite{paq2piq}, and \textbf{2)} aesthetic-related \textbf{other low-level attributes}~\cite{aadb,clipiaa} which are discernible to human perception and evoke varied emotions. Several studies~\cite{nima,paq2piq,atqa} follow this paradigm and categorize them through a relative golden standard, that whether the attributes \textit{directly improve or degrade picture quality} (\textit{Yes$\to$Distortions; No$\to$Others}).

\textbf{Axis 2: Global Perception \textit{vs} Local In-context Perception.} In recent research on low-level vision, it is observed that human perceptions of low-level visuals often intertwine with higher-level contextual comprehension~\cite{sfa,rfugc,wu2022fastervqa,qalign}. For instance, a {{clear sky} might lack complex textures yet display exceptional clarity}. Furthermore, localized low-level appearances can deviate from their overall counterparts, as observed by~\cite{fastvqa,pvq}. Acknowledging these differences, we curate \textbf{local in-context perception} (Fig.~\ref{fig:2} \textit{right top}) questions, that require MLLMs to grasp the content or other context to answer correctly, while other questions are categorized as \textbf{global perception} (Fig.~\ref{fig:2} \textit{left top}).

\textit{2) Low-level Visual Concerns for Image Pairs}

\textbf{Axis 1: Distortions \textit{vs} Other Low-level Attributes.} 
Same as Axis 1 for single images. Please refer to Sec.~\ref{sec:221} \textit{1)}.

\textbf{Axis 2: Compare \textit{vs} Joint.} This dual approach mimics human visual perception more closely. Humans often use both comparison (looking at differences and similarities) and joint analysis (perceiving images in a unified context) when viewing images. The \textbf{comparison} highlights the differences and similarities between the two images, which is the key component of the full-reference IQA tasks~\cite{hore2010image} and other low-level enhancement evaluation tasks~\cite{zhang2021no,zhang2022no}. The \textbf{joint analysis}, on the other hand, looks at the images as a combined entity to understand the overall context or to detect patterns that emerge only when the images are considered together.

\textit{3) Question Types}
\label{sec:222}

In the \textbf{LLVisionQA$^+$} dataset, we curate three question types, \textit{Yes-or-No}, \textit{What}, and \textit{How} to simulate multiple query forms from humans. The details of the three question types are defined as follows.

\textbf{Type 1: \textit{Yes-or-No} Questions.} The fundamental type of questions is \textit{Yes-or-No}, \textit{i.e.}, judgments. 
Specifically, we notice that some MLLMs especially prefer to respond with \textit{yes} rather than \textit{no}. To reduce such biases in our benchmark, though designing questions with answers as \textit{yes} is easier, we ensure that around 40\% of all judgments are with correct answers as \textit{no}, via querying on \textbf{contrastive} low-level attributes or \textbf{non-existing} low-level attributes.

\textbf{Type 2: \textit{What} Questions.} Despite \textit{Yes-or-No} judgments, the \textit{what} questions are also a common type of queries in recent MLLM benchmarks such as~\cite{emabench}. In \textbf{Q-bench$^+$}, they classify low-level attributes in pictures (\textit{e.g., What distortion occurs in the image?}), or associated context given specific low-level appearances (for in-context perception questions, \textit{e.g., Which object in the image is under-exposed?}). Unlike \textit{Yes-or-No} questions, the \textit{What} questions examine more comprehensive low-level attribute understanding of MLLMs, by requiring correct perception on \textbf{multiple} attributes.

\textbf{Type 3: \textit{How} Questions.} Despite the two common types, we also include a special type, the \textit{How} questions, to cover non-extreme appearances~\cite{wu2023explainable} of low-level attribute dimensions into our benchmark, as an extension to \textit{Yes-or-No} questions. As shown in Fig.~\ref{fig:2}, we can query \textit{How is the clarity of the image?} for the image with both clear and blurry areas, and answer with \textbf{Medium}. With this special question type, we broaden the \textbf{Q-bench$^+$} into \textbf{finer-grained} low-level perception.

\begin{figure*}
    \centering
    \includegraphics[width=0.92\linewidth]{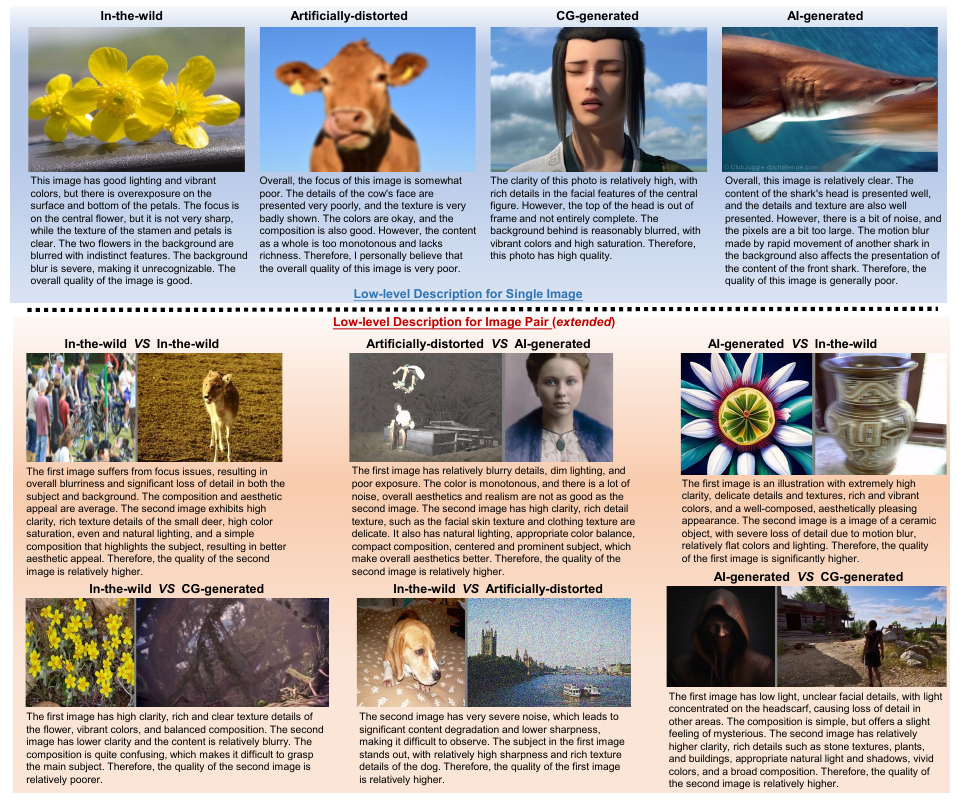}
    \vspace{-8pt}
    \caption{A dataset card of \textbf{LLDescribe$^+$} that evaluates the low-level \textbf{description} ability of MLLMs. 499 single images and 450 image pairs from 10 diverse sources are labeled with \textit{golden} descriptions, to serve as \textbf{\underline{text}} references to evaluate the completeness, preciseness, and relevance of MLLM outputs.}
    \label{fig:3}
    \vspace{-10pt}
\end{figure*}

\textit{4) GPT-assisted Evaluation Process}
\label{sec:223}

The input query format for MLLMs is as follows: 
\begin{itemize}
    \item \noindent \textbf{Single Images:} \\
\noindent \textit{{\small \#User: How is the clarity of the image? {\tt(Question)}\\ {[IMAGE\_TOKEN]} {\tt(Image)} \\ Choose between one of the following options:\\  A. High {\tt{(Correct)}}\t B. Medium{\tt(Wrong)}\t  C. Low{\tt(Wrong)}\\ \#Assistant:}}
    \item \noindent \textbf{Image Pairs:} \\
\noindent \textit{{\small \#User: Which image is brighter? {\tt(Question)}\\ The first image: {[IMAGE\_TOKEN]} {\tt(Image 1)} \\
The second image: {[IMAGE\_TOKEN]} {\tt(Image 2)} \\ Choose between one of the following options:\\  \t A. The first image{\tt{(Wrong)}} B. The second image{\tt(Correct)}\\ \#Assistant:}}
\end{itemize}
The correct answer has been shuffled and finally uniformly distributed among all choices (A/B/C/D). Moreover, while traditional visual question answering~\cite{cocovqa,okvqa} tasks typically employ traditional language metrics (BLEU-4, CIDEr) to compare performance, as observed by recent studies~\cite{mplugowl} and validated by us, most MLLMs cannot consistently provide outputs on \textbf{instructed formats}. Given the question above, different MLLMs may reply \textit{``A.''}, \textit{``High''}, \textit{``The clarity of the image is high."}, \textit{``The image is of high clarity."} (all correct), which are difficult to be exhaustively-included under traditional metrics. To solve this problem, we design, validate, and employ a \textbf{5-round} GPT-assisted evaluation process inspired by~\cite{mmbench}. Under this process, the question, correct answers, and MLLM replies are fed into GPT for evaluation.

\subsection{Benchmark on Low-level \textbf{Description} Ability}

In the second task of \textbf{Q-Bench$^+$}, we evaluate the language \textbf{description} ability of MLLMs on low-level information. This task is a sibling task of image captioning~\cite{cococaps,flickrcaps,nocaps} that describes image content with natural language, with a specific concern on the low-level appearance of images. To evaluate this ability automatically, we first derive a \textit{golden} low-level description dataset, denoted as \textbf{LLDescribe$^+$} (Sec.~\ref{sec:231}), including one long (\textit{average 58 words}) \textit{golden} description provided by experts for each of 499 images. With these \textit{golden} text descriptions, we are able to measure the quality of output low-level descriptions from MLLMs with a single-modal GPT, under the three dimensions: \textbf{completeness}, \textbf{preciseness}, as well as \textbf{relevance} (Sec~\ref{sec:232}). The discussions of the \textit{golden} descriptions and the evaluation process are as follows.

\subsubsection{Defining \textit{Golden} Low-level Descriptions for Images}
\label{sec:231}

For the description ability, MLLMs should accurately and completely describe low-level visual information of images. Thus, the \textit{ground truths} for these MLLMs are also built within a basic principle to cover as many low-level concerns as possible, so long as they are enumerated in Sec.~\ref{sec:221} and occur in images. The resulting \textit{golden} descriptions in \textbf{LLDescribe$^+$} have an average duration of \textbf{58} words, notably longer than common high-level image caption datasets (\textbf{11} for~\cite{nocaps}, \textbf{10} for~\cite{cococaps}).
Similar to the \textbf{LLVisionQA$^+$} dataset for the perception task, the 499 single images and 450 image pairs in \textbf{LLDescribe$^+$} dataset also include all 10 sources (as in Table~\ref{tab:1}) to cover images with diverse low-level appearances. The \textit{golden} descriptions on different sources of images are depicted in Fig.~\ref{fig:3}.

\subsubsection{Evaluation with Single-modal GPT}
After collecting the \textit{golden} descriptions, we design an input prompt to acquire the output descriptions from MLLMs:
\begin{itemize}
    \item \noindent \textbf{Single Images:} \\
\noindent \textit{{\small \#User: Describe the quality, aesthetics and other low-level appearance of the image in details. {\tt(Prompt)}\\ {[IMAGE\_TOKEN]} {\tt(Image)}\\ \#Assistant:}}

    \item \noindent \textbf{Image Pairs:} \\
\noindent \textit{{\small \#User: Compare and jointly analyze the quality, aesthetics and other low-level appearance of the images in details. {\tt(Prompt)}\\ The first image: {[IMAGE\_TOKEN]} {\tt(Image 1)} \\
The second image: {[IMAGE\_TOKEN]} {\tt(Image 2)} \\ \#Assistant:}}
\end{itemize}
Recent studies~\cite{vicuna} have proved single-modal GPT~\cite{openai2023gpt4} to be a reliable evaluation tool for pure language tasks. Via the \textbf{LLDescribe$^+$} dataset, we convert the multi-modality problem into a text-only setting, by matching the MLLM outputs with the \textit{golden} descriptions with single-modal GPT under three dimensions: \textbf{(1) Completeness.} More matched information with the \textit{golden} description is encouraged. \textbf{(2) Preciseness.} The controversial information with the \textit{golden} description is punished. \textbf{(3) Relevance.} More proportions of MLLM outputs should be related to low-level information, instead of others. Each dimension is scored among [0,1,2]. Similar as Sec.~\ref{sec:223}, we repeat \textbf{5 rounds} for each single evaluation and collect the weighted average as the final score. 

\label{sec:232}

\subsection{Benchmark on Precise Quality \textbf{Assessment} Ability}

\begin{figure*}[!t]
    \centering
    \includegraphics[width=0.9\linewidth]{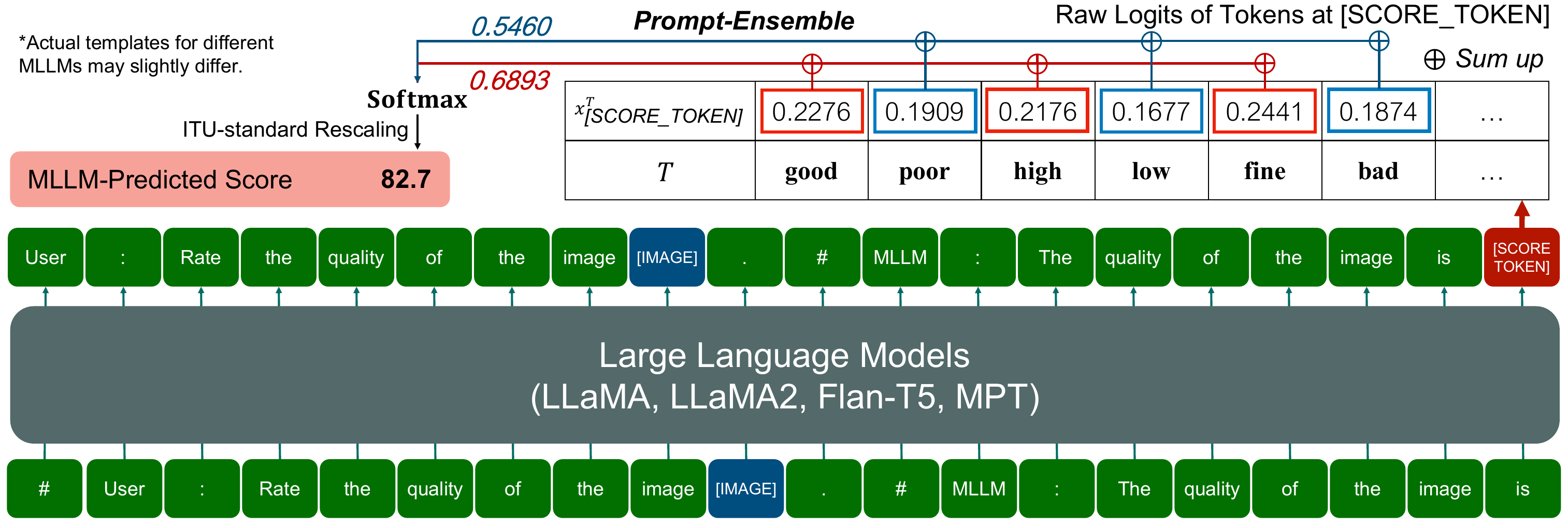}
    \vspace{-10pt}
    \caption{The proposed softmax-based strategy. Instead of directly decoding tokens from the \textit{[SCORE\_TOKEN] position}, the strategy extracts log probabilities (logits) of \textbf{\textit{positve}} and \textbf{\textit{negative}} words (\textbf{\textit{good}} and \textbf{\textit{poor}} as default), and predicts \textit{quantifiable} score via a {\tt softmax} pooling between the two logits.}
    \label{fig:4}
    \vspace{-10pt}
\end{figure*}

\begin{table*}\small
    \centering
    \renewcommand\arraystretch{1.05}
    \renewcommand\tabcolsep{10pt}
    \caption{Results on the {\tt dev} and {\tt test} subsets of \textbf{LLVisionQA$^+$} for the low-level \textbf{Perception} ability of MLLMs. Open-source MLLMs with \textit{top-3} performance in each sub-category are marked with best in \textbf{bold} and second/third \underline{underlined}.}
    \vspace{-8pt}
    \resizebox{\linewidth}{!}{\begin{tabular}{l|ccc|cc|cc|c}
    \toprule
        \textbf{Sub-categories} & \multicolumn{3}{c|}{\textbf{Question Types}} & \multicolumn{4}{c|}{\textbf{Quadrants of Low-level Concerns}} & \multirow{3}{*}{{\textit{Overall$\uparrow$}}} \\ \cdashline{1-8}
        \multirow{2}{*}{\textbf{Model} \textit{(variant)}}  & \multirow{2}{*}{\textit{Yes-or-No$\uparrow$}}& \multirow{2}{*}{\textit{What$\uparrow$}} & \multirow{2}{*}{\textit{How$\uparrow$}} & \multirow{2}{*}{\textit{Distortion$\uparrow$}} & \multirow{2}{*}{\textit{Other$\uparrow$}} & \textit{In-context}  &\textit{In-context}  \\
        &&&&&&\textit{Distortion$\uparrow$}& \textit{Other$\uparrow$} \\ \hline
        \textit{\textbf{Dev Set /}}\textit{ random guess} & 50.00\% & 27.86\% & 33.31\% & 37.89\% & 38.48\% & 38.28\% & 35.82\% & 37.80\% \\ \cdashline{1-9}
    InfiMM (\textit{Zephyr-7B})~\cite{InfiMM} & 57.45\% & 57.96\% & 44.62\% & 47.27\% & 57.17\% & 49.67\% & 64.08\% & 53.37\% \\
    Emu2-Chat (\textit{LLaMA-33B})~\cite{emu2} & \underline{71.81}\% & \underline{67.25}\% & 56.18\% & \textbf{64.78}\% & 63.19\% & \underline{63.48}\% & 72.24\% & \underline{65.28}\% \\
    Fuyu-8B (\textit{Persimmon-8B})~\cite{fuyu-8b} &53.33\% & 43.70\% & 38.00\% & 40.81\% & 47.40\% & 45.45\% & 49.23\% & 45.05\% \\
    BakLLava (\textit{Mistral-7B})~\cite{bakllava}  & 66.00\% & 56.16\% & 51.12\% & 51.15\% & 61.57\% & 53.72\% & 72.00\% & 57.48\% \\
    SPHINX~\cite{sphinx}  & \textbf{74.18}\% & \textbf{68.81}\% & \textbf{62.07}\% & \underline{63.62}\% & \textbf{71.76}\% & \textbf{66.12}\% & \textbf{76.33}\% & \textbf{68.56}\% \\
     mPLUG-Owl2 \textit{(LLaMA-7B)}~\cite{mplug2} &  \underline{72.18}\% & 57.96\% & 56.19\% & 56.68\% & \underline{69.21}\%& 53.29\%& 72.65\%& 61.61\% \\
    LLaVA-v1.5 (\textit{Vicuna-v1.5-7B})~\cite{improvedllava}  & 66.36\% & 58.19\% & 50.51\% & 49.42\% & {65.74}\% & 54.61\% & {70.61}\% & 58.66\% \\
    LLaVA-v1.5 (\textit{Vicuna-v1.5-13B})~\cite{improvedllava} & 65.27\% & {64.38}\% & \underline{56.59}\% & 56.03\% & {67.13}\% & {61.18}\% & 67.35\% & {62.14}\% \\
    InternLM-XComposer-VL \textit{(InternLM)}~\cite{xcomposer} & {69.45}\% & \underline{65.27}\% & \underline{60.85}\% & \underline{61.67}\% & \underline{70.14}\% & 56.91\% & \underline{75.10}\% & \underline{65.35}\% \\
    IDEFICS-Instruct   \textit{(LLaMA-7B)}~\cite{idefics} & 56.18\% & 44.69\% & 44.02\% & 42.80\% & 54.17\% & 44.74\% & 56.33\% & 48.70\% \\
    Qwen-VL \textit{(QwenLM)}~\cite{Qwen-VL} & 63.09\% & 58.19\% & {56.39}\% & 50.58\% & 62.73\% & 57.89\% & \underline{73.88}\% & 59.40\% \\
    Shikra \textit{(Vicuna-7B)}~\cite{shikra} & 65.64\% & 47.35\% & 49.09\% & 48.83\% & 59.49\% & 50.00\% & 64.08\% & 54.65\% \\
    Otter-v1   \textit{(MPT-7B)}~\cite{otter} & 57.09\% & 40.71\% & 39.55\% & 42.22\% & 49.31\% & 44.08\% & 52.65\% & 46.35\% \\
    InstructBLIP  \textit{(Flan-T5-XL)}~\cite{iblip} & {67.64}\% & {59.96}\% & 55.98\% & {56.23}\% & 65.51\% & {58.22}\% & 69.39\% & {61.47}\% \\
    InstructBLIP   \textit{(Vicuna-7B)}~\cite{iblip} & {71.64}\% & 52.65\% & 43.81\% & 48.64\% & 62.50\% & 55.59\% & 64.90\% & 56.72\% \\
    VisualGLM-6B   \textit{(GLM-6B)}~\cite{glm} & 60.18\% & 54.20\% & 46.25\% & 51.75\% & 54.40\% & 53.62\% & 57.14\% & 53.78\% \\
    mPLUG-Owl  \textit{(LLaMA-7B)}~\cite{mplugowl} & 66.00\% & 54.87\% & 44.02\% & 51.36\% & 55.09\% & 54.28\% & 65.71\% & 55.38\% \\
    LLaMA-Adapter-V2~\cite{llamaadapterv2} & 66.18\% & 59.29\% & 52.13\% & {57.39}\% & 56.25\% & \underline{63.16}\% & 64.90\% & 59.46\% \\
    LLaVA-v1 (\textit{Vicuna-13B})~\cite{llava} & 54.00\% & 53.10\% & 55.38\% & 48.64\% & 54.63\% & 55.59\% & 63.27\% & 54.18\% \\
    MiniGPT-4  \textit{(Vicuna-13B)}~\cite{minigpt4} & 55.82\% & 50.22\% & 40.37\% & 42.02\% & 48.38\% & 51.97\% & 61.22\% & 49.03\% \\

    \hdashline
    \textbf{Qwen-VL-Plus} (\textit{Closed-source})~\cite{Qwen-VL}  & 73.77\% & 69.47\% & 53.88\% & 66.21\% & 65.72\% & 63.81\% & 68.75\% & 66.04\%  \\
    \textbf{Qwen-VL-Max} (\textit{Closed-source})~\cite{Qwen-VL}  & 75.60\% & 79.43\% & 66.09\% & 73.39\% & 74.08\% & 71.0\% & 76.92\% & 73.63\% \\
    \textbf{Gemini-Pro} (\textit{Closed-source})~\cite{geminipro}  & 68.80\% & 73.74\% & 62.34\% & 66.30\% & 71.34\% & 63.91\% & 73.09\% & 68.16\%  \\ 
    \textbf{GPT-4V} (\textit{Closed-source})~\cite{openai2023gpt4} & 76.85\% & 79.17\% & 67.52\% & 73.53\% & 76.18\% & 72.83\% & 76.47\% & 74.51\% \\ 
        \midrule
    \textit{\textbf{Test Set /}}\textit{ random guess} & 50.00\% & 28.48\% & 33.30\% & 37.24\% & 38.50\% & 39.13\% & 37.10\% & 37.94\% \\ \cdashline{1-9}
         InfiMM (\textit{Zephyr-7B})~\cite{InfiMM} & 61.31\% & 56.61\% & 49.58\% & 47.79\% & 62.05\% & 51.71\% & 67.68\% & 56.05\%\\
         Emu2-Chat (\textit{LLaMA-33B})~\cite{emu2}  & 70.09\% & \underline{65.12}\% & 54.11\% & \textbf{66.22}\% & 62.96\% & \underline{63.47}\% & 73.21\% & \underline{64.32}\% \\
         Fuyu-8B (\textit{Persimmon-8B})~\cite{fuyu-8b} & 62.22\% & 35.79\% & 36.62\% & 41.07\% & 49.40\% & 45.89\% & 49.04\% & 45.75\% \\
         BakLLava (\textit{Mistral-7B})~\cite{bakllava} & 66.46\% & 61.48\% & 54.83\% & 51.33\% & 63.76\% & 56.52\% & \textbf{78.16}\% & 61.02\% \\
         SPHINX~\cite{sphinx} & \textbf{74.45}\% & \textbf{65.50}\% & \textbf{62.13}\% & \underline{59.11}\% & \textbf{73.26}\% & \textbf{66.09}\% & \underline{77.56}\% & \textbf{67.69}\% \\
         mPLUG-Owl2 \textit{(LLaMA-7B)}~\cite{mplug2} & \underline{72.26}\% & 55.53\% & \underline{58.64}\% & 52.59\% & \underline{71.36}\% & 58.90\% & 73.00\% & 62.68\% \\
        LLaVA-v1.5 (\textit{Vicuna-v1.5-7B})~\cite{improvedllava} & 64.60\% & 59.22\% & 55.76\% & 47.98\% & {67.30}\% & {58.90}\% & {73.76}\% & 60.07\% \\
        LLaVA-v1.5 (\textit{Vicuna-v1.5-13B})~\cite{improvedllava} & 64.96\% & \underline{64.86}\% & 54.12\% & 53.55\% & {66.59}\% & {58.90}\% & 71.48\% & 61.40\% \\
        InternLM-XComposer-VL \textit{(InternLM)}~\cite{xcomposer} & 68.43\% & {62.04}\% & \underline{61.93}\% & {56.81}\% & \underline{70.41}\% & 57.53\% & \underline{77.19}\% & \underline{64.35}\% \\
        IDEFICS-Instruct  \textit{(LLaMA-7B)}~\cite{idefics} & 60.04\% & 46.42\% & 46.71\% & 40.38\% & 59.90\% & 47.26\% & 64.77\% & 51.51\% \\
        Qwen-VL \textit{(QwenLM)}~\cite{Qwen-VL} & 65.33\% & {60.74}\% & {58.44}\% & 54.13\% & 66.35\% & 58.22\% & {73.00}\% & {61.67}\% \\
        Shikra (\textit{Vicuna-7B})~\cite{shikra} & 69.09\% & 47.93\% & 46.71\% & 47.31\% & 60.86\% & 53.08\% & 64.77\% & 55.32\% \\
        Otter-v1 \textit{(MPT-7B)}~\cite{otter} & 57.66\% & 39.70\% & 42.59\% & 42.12\% & 48.93\% & 47.60\% & 54.17\% & 47.22\% \\
        InstructBLIP \textit{(Flan-T5-XL)}~\cite{iblip} & {69.53}\% & 59.00\% & {56.17}\% & \underline{57.31}\% & 65.63\% & 56.51\% & 71.21\% & {61.94}\% \\
        InstructBLIP \textit{(Vicuna-7B)}~\cite{iblip} & {70.99}\% & 51.41\% & 43.00\% & 45.00\% & 63.01\% & 57.19\% & 64.39\% & 55.85\% \\
        VisualGLM-6B \textit{(GLM-6B)}~\cite{glm} & 61.31\% & 53.58\% & 44.03\% & 48.56\% & 54.89\% & 55.48\% & 57.79\% & 53.31\% \\
        mPLUG-Owl  \textit{(LLaMA-7B)}~\cite{mplugowl} & \underline{72.45}\% & 54.88\% & 47.53\% & 49.62\% & 63.01\% & \underline{62.67}\% & 66.67\% & 58.93\% \\
        LLaMA-Adapter-V2~\cite{llamaadapterv2} & 66.61\% & 54.66\% & 51.65\% & {56.15}\% & 61.81\% & {59.25}\% & 54.55\% & 58.06\% \\
        LLaVA-v1 (\textit{Vicuna-13B})~\cite{llava} & 57.12\% & 54.88\% & 51.85\% & 45.58\% & 58.00\% & 57.19\% & 64.77\% & 54.72\% \\
        MiniGPT-4 \textit{(Vicuna-13B)}~\cite{minigpt4} & 60.77\% & 50.33\% & 43.00\% & 45.58\% & 52.51\% & 53.42\% & 60.98\% & 51.77\% \\
        \hdashline
        \textbf{Qwen-VL-Plus} (\textit{Closed-source})~\cite{Qwen-VL}  & 75.74\% & 73.25\% & 57.33\% & 64.88\% & 73.24\% & 68.67\% & 70.56\% & 68.93\% \\
        \textbf{Qwen-VL-Max} (\textit{Closed-source})~\cite{Qwen-VL}  & 73.20\% & 81.02\% & 68.39\% & 70.84\% & 74.57\% & 73.11\% & 80.44\% & 73.90\% \\
        \textbf{Gemini-Pro} (\textit{Closed-source})~\cite{geminipro}  & 71.26\% & 71.39\% & 65.59\% & 67.30\% & 73.04\% & 65.88\% & 73.60\% & 69.46\% \\ 
         \textbf{GPT-4V} (\textit{Closed-source})~\cite{openai2023gpt4} & 77.72\% & 78.39\% & 66.45\% & 71.01\% & 71.07\% & 79.36\% & 78.91\% & 74.10\%  \\ \hdashline
         \textit{Junior-level \textit{Human}} &82.48\% & 79.39\% & 60.29\% & 75.62\% & 72.08\% & 76.37\% & 73.00\% & 74.31\%  \\
        \textit{Senior-level \textit{Human}} &84.31\% & 88.94\% & 72.02\% & 79.65\% & 79.47\% & 83.90\% & 87.07\% & 81.74\%  \\ \bottomrule
   \end{tabular}}
    \vspace{-12pt}
    \label{tab:perception}
\end{table*}

In the third task, we benchmark the ability of MLLMs to provide \textit{quantifiable} \textbf{assessment} on the overall low-level appearance of images. Unlike the two tasks above, we utilize existing IQA datasets that are collected across a variety of low-level appearances to evaluate how MLLMs can predict \textit{quantifiable} quality scores {aligned with human opinions}. All the three types of IQA datasets (\textit{in-the-wild}, \textit{generated}, \textit{artificially-distorted}) as mentioned in Sec.~\ref{sec:21} are evaluated, to provide a broad range measurement of the assessment ability of MLLMs. Nevertheless, how to collect \textit{quantifiable} quality scores from MLLMs remains challenging as their outputs only have weak measurability (Sec.~\ref{sec:241}). Noticing that MLLMs can provide probabilities of tokens, we employ {\tt softmax} pooling on the logits of \textbf{\textit{good}} and \textbf{\textit{poor}} under a simple and direct prompt template, deriving into \textit{quantifiable} quality scores (Sec.~\ref{sec:242}), as illustrated in Fig.~\ref{fig:4}. Details as follows.

\subsubsection{Weak Measurability of MLLM Outputs}
\label{sec:241}

In \textbf{Q-Bench$^+$}, we aim to fairly compare the \textbf{assessment} ability between different MLLMs on diverse low-level appearances. Henceforth, our principle is to define a unified, simplest instruction that is applicable for all MLLMs on all IQA datasets. Under this principle, we conduct toy experiments on Shikra~\cite{shikra} and LLaVA-v1~\cite{llava}, with two simple instruction strategies: \textbf{(A) Direct Instruction,} in which the prompt is designed as simple as \textit{``Rate the quality of the image''}. The top-frequency answers are  \textbf{\textit{good}} (78\%), and \textbf{\textit{poor}} (20\%), with other outputs almost negligible. \textbf{(B) Numerical Instruction,} in which we specifically instruct numerical ratings, with the prompt: \textit{``Score the quality of the image from {1 to 5}, with 1 as lowest and 5 as highest.''}. Under the numerical strategy, the top-frequency answers are \textbf{5} (84\%), \textbf{1} ({9\%}), and \textbf{3} (5\%); though within the score range, the frequencies of scores \textbf{2} and \textbf{4} are both less than 1\%. The toy experiments imply the weak measurability of MLLM outputs, given that the answers are statistically \textbf{1)} biased towards \textit{positive}, \textbf{2)} biased towards \textit{extreme}, and \textbf{3)} with \textit{only two} effective scales. Therefore, it is necessary to explore extended strategies for MLLMs to provide truly \textit{quantifiable} outputs for low-level \textbf{assessment}.

\subsubsection{A Softmax-based Evaluation Strategy}
\label{sec:242}

Given the above observations, we design the softmax-based evaluation strategy (Fig.~\ref{fig:4}) to reduce the negative impacts of the biases and lack of scales. To start with, we design our strategy within the \textbf{Direct Instruction}, which is more general and less biased than the \textbf{Numerical Instruction}. The strategy is based on the observation that two top-frequency outputs, \textbf{\textit{good}} and \textbf{\textit{poor}}, can be considered as anchors for better and worse human perception, and the \textbf{Direct Strategy} can be approximated into a binary classification problem on the \textit{[SCORE\_TOKEN]} position, or technically, an {\tt argmax} between the logits of \textbf{\textit{good}} ($x^\text{\textbf{good}}_{\textit{SCORE\_TOKEN}}$) and \textbf{\textit{poor}} ($x^\text{\textbf{poor}}_{\textit{SCORE\_TOKEN}}$) on this position.

{ Specifically, the full input prompt for obtaining quality scores is formatted like this:
\begin{itemize}
    \item \noindent \textbf{Assessment Prompt:} \\
    \noindent \textit{{\small \#User: Rate the quality of the image \{[IMAGE\_TOKEN]\} {\tt(Image)}. 
    The quality of the image is} {\tt(Prompt)}\\ {\#Assistant: \{[SCORE\_TOKEN]\}. {\tt(Response)} }}
\end{itemize}
As seen from the prompt, we strictly adhere to the structure '\textit{Rate the quality of the image {\tt(Image).} The quality of the image is}' and obtain only the descriptive response (\textit{[SCORE\_TOKEN]}) from the MLLM. The (\textit{[SCORE\_TOKEN]}) can be interpreted as a probability map that includes the log probabilities for all words in the vocabulary, encompassing quality-related terms such as \textit{good} and \textit{poor}. }
In our revised strategy, we modify the {\tt argmax} into {\tt softmax} to collect better \textit{quantifiable} scores:
\begin{equation}
q_\mathrm{pred} = \frac{e^{x^\text{\textbf{good}}_{\textit{SCORE\_TOKEN}}}}{e^{x^\text{\textbf{good}}_{\textit{SCORE\_TOKEN}}}+e^{x^\text{\textbf{poor}}_{\textit{SCORE\_TOKEN}}}}
\label{eq:1}
\end{equation}
This simple and generally-applicable strategy enables us to collect \textit{quantifiable} outputs ($q_\mathrm{pred}$) from MLLMs with higher correlation to human ratings, as verified in our experimental analysis (Fig.~\ref{fig:softmax}).

\subsubsection{Prompt Ensemble for Boosting Quantitative Abilities of MLLMs} { Multiple synonym prompts can broaden the semantic range, allowing for a more nuanced understanding that might be missed by a single term. Additionally, multiple synonym prompts diminish uncertainty since diverse terms have subtly different meanings, resulting in a more dependable assessment. Specifically, we further choose the combination prompts of [\textbf{\textit{good}}, \textbf{\textit{high}}, \textbf{\textit{fine}}] and [\textbf{\textit{poor}}, \textbf{\textit{low}}, \textbf{\textit{bad}}] to replace \textbf{\textit{good}} and \textbf{\textit{poor}} respectively. } The \textit{quantifiable} outputs ($q_\mathrm{pred}$) can then be altered as: 
\begin{equation}
q_\mathrm{pred} = \frac{e^{\sum_t^{t\in \mathcal{P}} x^t_{\textit{SCORE\_TOKEN}}}}{e^{\sum_t^{t\in \mathcal{P}} x^t_{\textit{SCORE\_TOKEN}}}+e^{\sum_t^{t\in \mathcal{N}} x^t_{\textit{SCORE\_TOKEN}}}}
\label{eq:ensemble}
\end{equation}
where $\mathcal{P}$ indicates the positive token set (from \textit{good}, \textit{fine}, \textit{high}, etc.), while $\mathcal{N}$ represents the negative token set (from \textit{poor}, \textit{bad}, \textit{low}, etc.).
\textbf{The implementation of the prompt ensemble approach does not add extra computational complexity.} The core computation occurs once the input prompt is entered and the language model generates the \textit{[SCORE\_TOKEN]}. After this, we only require tokenization of the words used, followed by the calculation of logits for the \textit{[SCORE\_TOKEN]}. The boosted performance is listed in Fig.~\ref{fig:ensemble}.

\begin{table*}\small
    \centering
    \renewcommand\arraystretch{1.05}
    \renewcommand\tabcolsep{12pt}
    \caption{Results on the {\tt dev} and {\tt test} subsets of \textbf{LLVisionQA$^+$} for the low-level \textbf{Perception-Pair} ability of MLLMs. TOpen-source MLLMs with \textit{top-3} performance in each sub-category are marked with best in \textbf{bold} and second/third \underline{underlined}.}
    \vspace{-8pt}
    \resizebox{\linewidth}{!}{\begin{tabular}{l|ccc|cc|cc|c}
    \toprule
        \textbf{Sub-categories} & \multicolumn{3}{c|}{\textbf{Question Types}} & \multicolumn{2}{c|}{\textbf{Low-level Concerns}} & \multicolumn{2}{c|}{\textbf{Pairwise Concerns}} & \multirow{3}{*}{{\textit{Overall$\uparrow$}}} \\ \cdashline{1-8}
        \multirow{2}{*}{\textbf{Model} \textit{(variant)}}  & \multirow{2}{*}{\textit{Yes-or-No$\uparrow$}}& \multirow{2}{*}{\textit{What$\uparrow$}} & \multirow{2}{*}{\textit{How$\uparrow$}} & \multirow{2}{*}{\textit{Distortion$\uparrow$}} & \multirow{2}{*}{\textit{Other$\uparrow$}} & \multirow{2}{*}{\textit{Compare$\uparrow$}}  &\multirow{2}{*}{\textit{Joint$\uparrow$}}  \\
        &&&&&&& \\ \hline
        \textit{\textbf{Dev Set /}}\textit{ random guess} & 50.00\% & 32.16\% & 33.30\% & 38.59\% & 41.74\% & 38.66\% & 43.89\% & 39.60\% \\ \cdashline{1-9}
        InfiMM (\textit{Zephyr-7B})~\cite{InfiMM} & 48.11\% & 39.04\% & 40.06\% & 42.56\% & 43.78\% & 41.77\% & 48.33\% & 42.95\% \\
        Emu2-Chat (\textit{LLaMA-33B})~\cite{emu2}  & 56.64\% & 41.15\% & \underline{49.62\%} & \underline{49.12\%}
        & {51.91\%} & 47.86\% & \underline{60.00\%} & \underline{50.05}\% \\
        Fuyu-8B (\textit{Persimmon-8B})~\cite{fuyu-8b} & \textbf{68.76\%} & 33.56\% & 38.78\% & {46.83\%} & \underline{54.03\%} & 47.86\% & 55.00\% & 49.15\%\\
        BakLLava (\textit{Mistral-7B})~\cite{bakllava} & 56.92\% & \textbf{43.83\%} & \textbf{50.00\%} & \underline{49.33\%} & \textbf{54.34\%} & \textbf{50.66\%} & 52.22\% & \textbf{50.94\%}\\
        mPLUG-Owl2 (\textit{Q-Instruct})~\cite{q-instruct} & \underline{59.19\%} & 42.12\% & 47.43\% & \textbf{49.63\%} & 52.48\% & \underline{49.81\%} & 53.88\% & \underline{50.54}\%\\
        mPLUG-Owl2 (\textit{LLaMA-7B})~\cite{mplug2} & 58.43\% & 39.72\% & \underline{48.39\%} & 49.04\% & 51.55\% & 47.50\% & \textbf{60.55\%} & {49.85\%} \\
        LLaVA-v1.5 (\textit{Vicuna-v1.5-7B})~\cite{improvedllava} & \underline{60.46\%} & \underline{42.85\%} & 41.53\% & {47.88\%} & {51.89\%} & {46.55\%} & \underline{59.57\%} & 49.32\% \\
        LLaVA-v1.5 (\textit{Vicuna-v1.5-13B})~\cite{improvedllava} & 56.42\% & \underline{42.46\%} & 48.38\% & 48.15\% & \underline{53.41\%} & \underline{48.84}\% & 54.44\% & {49.85\%} \\
        \hdashline
        \textbf{Qwen-VL-Plus} (\textit{Closed-source})~\cite{Qwen-VL}  & 63.63\% & 55.55\% & 55.71\% & 61.61\% & 56.52\% & 65.81\% & 58.45\% & 60.70\%  \\
        \textbf{Qwen-VL-Max} (\textit{Closed-source})~\cite{Qwen-VL}  & 71.96\% & 62.87\% & 65.53\% & 69.21\% & 62.69\% & 67.54\% & 66.01\% & 67.27\% \\
        \textbf{Gemini-Pro} (\textit{Closed-source})~\cite{geminipro} & 64.98\% & 51.36\% & 54.16\% & 58.17\% & 56.52\% & 57.73\% & 57.22\% & 57.64\%  \\ 
        \textbf{GPT-4V} (\textit{Closed-source})~\cite{openai2023gpt4} & 79.34\% & 70.54\% & 78.52\% & 75.84\% & 77.95\% & 78.80\% & 66.11\% & 76.52\% \\ 
        \midrule
        \textit{\textbf{Test Set /}}\textit{ random guess} & 50.00\% & 32.03\% & 33.16\% & 38.95\% & 41.95\% & 38.69\% & 43.70\% & 39.82\% \\ \cdashline{1-9}
        InfiMM (\textit{Zephyr-7B})~\cite{InfiMM} & 54.21\% & 43.38\% & 45.32\% & \underline{49.57}\% & 45.67\% & 48.32\% & 48.88\% & 48.44\% \\
        Emu2-Chat (\textit{LLaMA-33B})~\cite{emu2} &51.94\% & 29.78\% & \textbf{53.84\%} & 42.01\% & 55.71\% & 46.26\% & 49.09\% & 47.08\% \\
        Fuyu-8B (\textit{Persimmon-8B})~\cite{fuyu-8b} & \textbf{70.36\%} & 28.13\% & 35.98\% & 44.08\% & 57.43\% & 47.02\% & 51.11\% & 47.94\%\\
        BakLLava (\textit{Mistral-7B})~\cite{bakllava} & 60.09\% & \underline{45.42\%} & \underline{50.86\%} & \textbf{53.09\%} & \underline{58.82\%} & \textbf{54.52\%} & \underline{55.55\%} & \underline{52.75\%}\\
        mPLUG-Owl2 (\textit{Q-Instruct})~\cite{q-instruct} & \underline{60.24\%} & \textbf{47.46\%} & 48.78\% & \underline{52.81\%} & 53.97\% & 51.42\% & \underline{59.11\%} & \textbf{53.15\%}\\
        mPLUG-Owl2 (\textit{LLaMA-7B})~\cite{mplug2} & 58.07\% & 36.61\% & 48.44\% & 47.74\% & 51.90\% & 45.73\% & \textbf{60.00\%} & 48.94\% \\
        LLaVA-v1.5 (\textit{Vicuna-v1.5-7B})~\cite{improvedllava} & \underline{60.72\%} & 42.37\% & \underline{50.17\%} & {49.15\%} & \textbf{59.86\%} & \underline{52.97\%} & 49.77\% & \underline{52.25\%} \\
        LLaVA-v1.5 (\textit{Vicuna-v1.5-13B})~\cite{improvedllava} & 57.34\% & \underline{47.45\%} & 49.13\% & 49.01\% & \underline{59.51\%} & \underline{52.06\%} & 52.00\% & 52.05\% \\
        \hdashline
        \textbf{Qwen-VL-Plus} (\textit{Closed-source})~\cite{Qwen-VL}  & 66.85\% & 55.79\% & 59.91\% & 62.46\% & 58.77\% & 62.17\% & 59.20\% & 61.48\%  \\
        \textbf{Qwen-VL-Max} (\textit{Closed-source})~\cite{Qwen-VL}  & 67.65\% & 67.56\% & 65.35\% & 69.09\% & 61.18\% & 68.65\% & 61.29\% & 66.99\% \\
        \textbf{Gemini-Pro} (\textit{Closed-source})~\cite{geminipro}  & 65.78\% & 56.61\% & 56.74\% & 60.42\% & 60.55\% & 60.46\% & 60.44\% & 60.46\% \\
         \textbf{GPT-4V} (\textit{Closed-source})~\cite{openai2023gpt4} & 79.75\% & 69.49\% & 84.42\% & 77.32\% & 79.93\% & 81.00\% & 68.00\% & 78.07\%  \\ 
        \hdashline
         \textit{Junior-level Human}  & 78.11\% & 77.04\% & 82.33\% & 78.17\% &  77.22\% & 80.26\% & 76.39\% & 80.12\%\\
        \textit{Senior-level \textit{Human}} &83.00\% & 84.81\% & 89.85\% & 83.13\% & 90.78\% & 86.55\% & 82.25\% & 85.48\% \\
       \bottomrule
    \end{tabular}}
    \vspace{-10pt}
    \label{tab:perception_pair}
\end{table*}

\section{Experiment}
In \textbf{Q-Bench$^+$}, we evaluate the performance of up to \textbf{20} up-to-date popular and competitive open-source as well as \textbf{4} closed-source commercial MLLMs under {\textbf{zero-shot}} settings.

\subsection{Findings on \textbf{Perception}}
\label{sec:32}

For a holistic examination of the \textbf{perception} ability of MLLMs, we evaluate the multi-choice correctness of MLLMs on different sub-categories of the \textbf{LLVision$^+$} dataset, which is equally divided as {\tt dev} (\textit{will be released}) and {\tt test} (\textit{will keep private}) subsets as shown in Table~\ref{tab:perception} and Table~\ref{tab:perception_pair} respectively. \textbf{Only the MLLMs that support multiple images input} are included for the \textbf{perception-pair} ability benchmark.

\subsubsection{Perception for Single Images}  a) We are glad that the majority of MLLMs can significantly outperform \textit{random guess} on all sub-categories as shown in Table~\ref{tab:perception}. Considering that all participating MLLMs are without any explicit training on low-level visual attributes, these results show strong potentials for these general-purpose models when further fine-tuned with respective low-level datasets. b) Among all open-source MLLMs, the recently-released SPHINX reaches the best accuracy on this {question-answering} task, followed by Emu2-Chat and InternLM-XComposer-VL, which show rather close results. By achieving \textbf{more than 64\%} accuracy on both subsets, these models show exciting potential as robust low-level visual assistants in the future.  c) { Another key observation is that almost all methods \textbf{perceive worse on distortions} than other low-level attributes, which indicates that distortion questions are relatively more challenging. Upon detailed reflection, we have identified two primary possible reasons: c-1) The multi-modal datasets employed for pre-training MLLMs primarily target high-level tasks like classification and identification. These datasets seldom include instructions on low-level vision patterns, such as blur and noise, which are essential for distortion-related tasks. Consequently, MLLMs are not well-optimized for recognizing distortions, leading to their subpar performance in handling such tasks. c-2) Distortions are particularly sensitive to resolutions, as highlighted by Wu et al.~\cite{wu2022fastvqa}. Yet, to accommodate computational constraints, most MLLMs utilize downsampling as a preprocessing step for images. This approach inevitably degrades the structural integrity of distortions, further complicating the MLLMs' ability to accurately perceive them. } d) \textbf{Closed-source MLLMs and Humans.}  It is widely acknowledged that commercial closed-source MLLMs are the leading models in various tasks. To evaluate the low-level \textbf{perception} abilities of these MLLMs, we gauge the accuracy of Qwen-VL-Plus (Alibaba), Qwen-VL-Max (Alibaba), Gemini-Pro (Google), and GPT-4V (OpenAI) on the subsets of \textbf{LLVision$^+$} dataset. All closed-source MLLMs achieve superior performance than all open-source MLLMs on the {\tt test} subset, which indicates that open-source MLLMs still fall behind on low-level visual ability. GPT-4V exhibits the most competitive performance and outperforms the best open-source MLLM (SPHINX) by a large margin (\textbf{+6\%}), and on par accuracy with the \textit{Junior-level Human}. Despite its prowess, there is still a way to go for GPT-4V before it can match the overall proficiency of the \textit{Senior-level Human (with experiences on low-level visual tasks}, \textbf{7\%} better than GPT-4V). Furthermore, across all categories, the results show that GPT-4V, much like its open-source counterparts, faces challenges in recognizing \textbf{distortions}.

\begin{table*}\small
    \centering
    \renewcommand\arraystretch{1.1}
    \renewcommand\tabcolsep{4.4pt}
        \caption{Results on the low-level \textbf{Description} ability of MLLMs. $P_i$ denotes frequency for score $i$.}
        \vspace{-8pt}
    \resizebox{\linewidth}{!}{\begin{tabular}{l|cccc|cccc|cccc|c}
    \toprule
        \textbf{Dimensions} & \multicolumn{4}{c|}{\textbf{Completeness}} & \multicolumn{4}{c|}{\textbf{Precision}} & \multicolumn{4}{c|}{\textbf{Relevance}} & \multirow{2}{*}{\textit{Sum.$\uparrow$}} \\ \cdashline{1-13}
        \textbf{Model} (\textit{variant}) & $P_0$ & $P_1$ & $P_2$ & \textit{score$\uparrow$}   &  $P_0$ & $P_1$ & $P_2$ & \textit{score$\uparrow$}   & $P_0$ & $P_1$ & $P_2$  & \textit{score$\uparrow$} \\ \hline
        InfiMM (\textit{Zephyr-7B})~\cite{InfiMM} & 29.61\% & 62.32\% & 7.77\% & 0.77 & 29.25\% & 31.90\% & 38.51\% & 1.08 & 2.16\% & 22.72\% & 74.58\% & 1.71 & 3.58 \\
        Emu2-Chat (\textit{LLaMA-33B})~\cite{emu2} & 20.01\% & 52.77\% & 27.22\% & \underline{1.07} & 24.66\% &  27.12\% & 48.22\% & {1.24} & 1.21\% & 9.91\% & 88.88\% & \textbf{1.88} & \underline{4.19} \\
        Fuyu-8B (\textit{Persimmon-8B})~\cite{fuyu-8b} & 25.54\% & 61.00\% & 13.46\% & 0.88 & {41.96}\% & {32.76}\% & 25.28\% & 0.83 & 2.99\% & 11.34\% & 85.67\% & 1.82 & 3.53 \\
        BakLLava (\textit{Mistral-7B})~\cite{bakllava} &24.31\% & 51.22\% & 24.47\% & 1.00 & {49.23}\% & 24.11\% & 26.66\% & 0.77 & 1.25\% & 36.22\% & {62.53}\% & {1.61} & 3.38 \\
        SPHINX~\cite{sphinx} & 27.96\% & 64.36\% & 7.33\% & 0.79 & 26.16\% & 32.42\% & 41.01\% & 1.14 & 1.69\% & 23.00\% & 74.61\% & 1.72 & 3.65 \\
        mPLUG-Owl2 (\textit{LLaMA-7B})~\cite{mplug2} & 27.71\% & 38.58\% & 33.71\% & 1.06 &28.11\% & 19.78\% & 52.11\% & 1.24 &7.91\% & 48.18\% & 43.91\% & 1.36 &  3.67 \\
        LLaVA-v1.5 (\textit{Vicuna-v1.5-7B})~\cite{improvedllava} & 27.48\% & 54.74\% & 17.78\% & 0.90 & 30.51\% & 26.04\% & 43.45\%  & 1.13 &  10.85\% & 60.34\% & 28.81\% & 1.18 & 3.21 \\
        LLaVA-v1.5 (\textit{Vicuna-v1.5-13B})~\cite{improvedllava} & 27.68\% & 53.78\% & 18.55\% & 0.91 & 25.45\% & 21.47\% & 53.08\% & \textbf{1.28} & 6.31\% & 58.75\% & 34.94\% & 1.29 & 3.47 \\
        InternLM-XComposer-VL \textit{(InternLM)}~\cite{xcomposer} & 19.94\% & 51.82\% & 28.24\% & \underline{1.08} & 22.59\% &  28.99\% & 48.42\% & \underline{1.26} & 1.05\% & 10.62\% & 88.32\% & \textbf{1.87} & \textbf{4.21} \\
        IDEFICS-Instruct \textit{(LLaMA-7B)}~\cite{idefics} & 28.91\% & 59.16\% & 11.93\% & 0.83 & 34.68\% & 27.86\% & 37.46\% & 1.03 & 3.90\% & 59.66\% & 36.44\% & 1.33 & 3.18 \\
        Qwen-VL \textit{(QwenLM)}~\cite{Qwen-VL} & 26.34\% & 49.13\% & 24.53\% & 0.98 & {50.62}\% & 23.44\% & 25.94\% & 0.75 & 0.73\% & 35.56\% & {63.72}\% & {1.63} & 3.36 \\
        Shikra (\textit{Vicuna-7B})~\cite{shikra} & 21.14\% & {68.33}\% & 10.52\% & 0.89 & 30.33\% & 28.30\% & 41.37\% & 1.11 & 1.14\% & {64.36}\% & 34.50\% & 1.33 & 3.34 \\
        Otter-v1 \textit{(MPT-7B)}~\cite{otter} & 22.38\% & 59.36\% & 18.25\% & 0.96 & {40.68}\% & {35.99}\% & 23.33\% & 0.83 & 1.95\% & 13.20\% & {84.85}\% & {1.83} & 3.61 \\
        Kosmos-2~\cite{kosmos2} & 8.76\% & {70.91}\% & 20.33\% & \textbf{1.12} & 29.45\% & {34.75}\% & 35.81\% & 1.06 & 0.16\% & 14.77\% & {85.06}\% & \underline{1.85} & \underline{4.03} \\
        InstructBLIP \textit{(Flan-T5-XL)}~\cite{iblip} & 23.16\% & 66.44\% & 10.40\% & 0.87 & 34.85\% & 26.03\% & 39.12\% & 1.04 & {14.71}\% & 59.87\% & 25.42\% & 1.11 & 3.02 \\
        InstructBLIP \textit{(Vicuna-7B)}~\cite{iblip} & 29.73\% & 61.47\% & 8.80\% & 0.79 & 27.84\% & 23.52\% & 48.65\% & 1.21 & {27.40}\% & 61.29\% & 11.31\% & 0.84 & 2.84 \\
        VisualGLM-6B \textit{(GLM-6B)}~\cite{glm} & {30.75}\% & 56.64\% & 12.61\% & 0.82 & {38.64}\% & 26.18\% & 35.18\% & 0.97 & 6.14\% & {67.15}\% & 26.71\% & 1.21 & 2.99 \\
        mPLUG-Owl \textit{(LLaMA-7B)}~\cite{mplugowl} & 28.28\% & 37.69\% & {34.03}\% & {1.06} & 26.75\% & 18.18\% & {55.07}\% & \textbf{1.28} & 3.03\% & 33.82\% & 63.15\% & 1.60 & {3.94} \\
        LLaMA-Adapter-V2~\cite{llamaadapterv2} & 30.44\% & 53.99\% & 15.57\% & 0.85 & 29.41\% & 25.79\% & 44.80\% & 1.15 & 1.50\% & 52.75\% & 45.75\% & 1.44 & 3.45 \\
        LLaVA-v1 (\textit{Vicuna-13B})~\cite{llava} & {34.10}\% & 40.52\% & {25.39}\% & 0.91 & 30.02\% & 15.15\% & {54.83}\% & 1.25 & 1.06\% & 38.03\% & 60.91\% & 1.60 & {3.76} \\
        MiniGPT-4 (\textit{Vicuna-13B})~\cite{minigpt4} & {34.01}\% & 32.15\% & {33.85}\% & {1.00} & 29.20\% & 15.27\% & {55.53}\% & \underline{1.26} & 6.88\% & 45.65\% & 47.48\% & 1.41 & 3.67 \\
         \bottomrule
    \end{tabular}}
    \vspace{-10pt} 
    \label{tab:description}
\end{table*}

\begin{table*}\small
    \centering
    \renewcommand\arraystretch{1.1}
    \renewcommand\tabcolsep{5.5pt}
        \caption{Results on the low-level \textbf{Description-Pair} ability of MLLMs. $P_i$ denotes frequency for score $i$.}
        \vspace{-8pt}
    \resizebox{\linewidth}{!}{\begin{tabular}{l|cccc|cccc|cccc|c}
    \toprule
        \textbf{Dimensions} & \multicolumn{4}{c|}{\textbf{Completeness}} & \multicolumn{4}{c|}{\textbf{Precision}} & \multicolumn{4}{c|}{\textbf{Relevance}} & \multirow{2}{*}{\textit{Sum.$\uparrow$}} \\ \cdashline{1-13}
        \textbf{Model} (\textit{variant}) & $P_0$ & $P_1$ & $P_2$ & \textit{score$\uparrow$}   &  $P_0$ & $P_1$ & $P_2$ & \textit{score$\uparrow$}   & $P_0$ & $P_1$ & $P_2$  & \textit{score$\uparrow$} \\ \hline
        InfiMM (\textit{Zephyr-7B})~\cite{InfiMM} & 30.75\% & 62.66\% & 6.22\% & 0.75 & 34.17\% & 38.84\% & 26.35\% & \underline{0.91} & 2.57\% & 30.84\% & 65.28\% & 1.61 & 3.28 \\
        Emu2-Chat (\textit{LLaMA-33B})~\cite{emu2} & 41.25\% & 54.33\% & 4.42\% & 0.63 & 38.11\% & 36.41\% & 25.48\% & {0.87} & 4.12\% & 38.61\% & 57.27\% & 1.53 & 3.03 \\
        Fuyu-8B (\textit{Persimmon-8B})~\cite{fuyu-8b} & 37.95\% & 52.17\% & 9.11\% & 0.70 & 37.68\% & 37.33\% & 23.73\% & 0.84 & 3.95\% & 31.15\% & 62.84\% & 1.56 & 3.12\\
        BakLLava (\textit{Mistral-7B})~\cite{bakllava} & 29.46\% & 59.77\% & 10.57\% & 0.80 & 40.0\% & 38.08\% & 21.33\% & 0.80 & 2.26\% & 15.06\% & 82.04\% & \underline{1.79} & 3.40\\
        mPLUG-Owl2 (\textit{Q-Instruct})~\cite{mplug2} & 15.25\% & 65.76\% & 18.32\% & \textbf{1.02} & 39.44\% & 40.18\% & 19.62\% & 0.79 & 0.09\% & 9.86\% & 89.02\% & \textbf{1.87} & \textbf{3.69}\\
         mPLUG-Owl2 \textit{(LLaMA-7B)}~\cite{mplug2} & 19.43\% & 65.54\% & 14.45\% & \underline{0.94} & 30.94\% & 43.71\% & 24.63\% & \textbf{0.92} & 3.79\% & 26.94\% & 68.28\% & \underline{1.63} & \underline{3.50}\\
        LLaVA-v1.5 (\textit{Vicuna-v1.5-7B})~\cite{improvedllava} & 19.68\% & 72.57\% & 7.19\% & 0.86 & 38.00\% & 40.04\% & 20.97\% & 0.82 & 2.13\% & 39.77\% & 56.66\% & 1.53 & 3.22 \\
        LLaVA-v1.5 (\textit{Vicuna-v1.5-13B})~\cite{improvedllava} & 18.77\% & 73.44\% & 7.79\% & \underline{0.89} & 34.66\% & 38.72\% & 26.62\% & \textbf{0.92} & 1.02\% & 34.59\% & 64.39\% & \underline{1.63} & \underline{3.44} \\
        \bottomrule
    \end{tabular}}
    \vspace{-10pt} 
    \label{tab:description_pair}
\end{table*}

\begin{table}[t]\small
    \centering
    \renewcommand\arraystretch{1.25}
    \renewcommand\tabcolsep{3pt}
        \caption{{ Results on the GPT reliability on assessing \textbf{Description} ability of MLLMs. Metrics are \textit{SRCC/PLCC} between GPT-predicted scores and human-rated scores. Rank is the rank order given by \textit{GPT / HUMANS}. The worst Avg. is marked in \textbf{BOLD}.}}
        \vspace{-8pt}
    \resizebox{\linewidth}{!}{\begin{tabular}{l|ccc|c|c}
    \toprule
        \textbf{Dimensions}/\textbf{Model} (\textit{variant}) & \textbf{Complete.} & \textbf{Precision} & \textbf{Relevance} & Avg.$\uparrow$ & \textbf{Rank}\\ \hline
        \multicolumn{5}{l}{\textit{Reliability performance for single image description}} \\ \hdashline
        Emu2-Chat (\textit{LLaMA-33B})~\cite{emu2} & 0.97/0.97 & 0.97/0.98 & 0.96/0.96 & 0.97/0.97 & 2 / 2\\
        InternLM-X. \textit{(InternLM)}~\cite{xcomposer} & 0.95/0.96 & 0.98/0.98 & 0.95/0.95 & 0.96/0.97 & 1 / 1\\
        Kosmos-2~\cite{kosmos2} & 0.96/0.96 & 0.95/0.95 & 0.96/0.96 & \textbf{0.95/0.96} & 3 / 3 \\ 
        mPLUG-Owl \textit{(LLaMA-7B)}~\cite{mplugowl} & 0.96/0.97 & 0.96/0.97 & 0.96/0.96 & 0.96/0.97 & 4 /  4\\
        LLaVA-v1 (\textit{Vicuna-13B})~\cite{llava} & 0.95/0.95 & 0.96/0.96 & 0.96/0.97 & 0.96/0.96 & 5 / 5 \\ \hline
        \multicolumn{5}{l}{\textit{Reliability performance for image pair description}} \\ \hdashline
        InfiMM (\textit{Zephyr-7B})~\cite{InfiMM} & 0.96/0.99 & 0.95/0.96 & 0.95/0.96 & 0.96/0.97 & 5 / 5\\
        BakLLava (\textit{Mistral-7B})~\cite{bakllava} & 0.96/0.97 & 0.96/0.96 & 0.97/0.97 & 0.96/0.97 & 4 / 4\\
        mPLUG-Owl2 (\textit{Q-Instruct})~\cite{mplug2} & 0.96/0.97 & 0.95/0.96 & 0.95/0.95 & 0.95/0.96 & 1 / 1\\
         mPLUG-Owl2 \textit{(LLaMA-7B)}~\cite{mplug2} & 0.96/0.97 & 0.97/0.98 & 0.95/0.95 & 0.96/0.97 & 2 / 2\\
        LLaVA-v1.5 (\textit{Vicuna-v1.5-13B})~\cite{improvedllava} & 0.96/0.96 & 0.95/0.96 & 0.96/0.97 & 0.96/0.96 & 3 / 3\\
         \bottomrule
    \end{tabular}}
    \vspace{-10pt} 
    \label{tab:reliable}
\end{table}

\subsubsection{Perception for Image Pairs} Perception for image pairs is far more difficult for MLLMs since this task not only requires MLLMs to have stable low-level visual capabilities, but also requires MLLMs to be able to analyze two images simultaneously and conduct discerning comparisons. To enrich the MLLM diversity, we further include the mPLUG-Owl2 fine-tuned with the single image low-level visual dataset \textbf{Q-Instruct} \cite{q-instruct} for comparison. The performance is exhibited in Table~\ref{tab:perception_pair}. With closer inspections, we can obtain several interesting findings. a) \textbf{Open-source MLLMs are poor low-level comparators.} It seems that although they might show strong performance for single image perception, they are quite confused by the image pairs. Most of them get worse performance on the \textbf{Compare} subset than the \textbf{Joint} subset, which further confirms this point. For mPLUG-Owl2 (\textit{Q-Instruct}), despite being fine-tuned with the single image low-level visual dataset \textbf{Q-Instruct} \cite{q-instruct}, the overall performance improvement from the low-level knowledge infusion of single images is relatively weak. This also suggests that there is a necessity to build open-source low-level datasets for multiple images to cultivate the corresponding capabilities of open-source MLLMs.  b) \textbf{Closed-source MLLMs are more robust in this task.} This may be because these closed-source MLLMs are supported by training on multiple-image data, allowing them to make better comparative judgments. Particularly with GPT-4V, its performance in the \textbf{compare} subset is significantly higher than in the \textbf{joint} subset, and it far exceeds all other models, even reaching the level of a junior human. c) \textbf{Perception for image pairs is easier for humans.} Comparing image pairs is simpler for humans, as the answers to related questions tend to be more objective. Especially for junior-level humans with no professional experience, they may have stronger subjectivity in grasping absolute sensations, but it is easier to remain objective when they are faced with comparison concepts. For example, it's difficult for a junior-level human to judge whether the lighting in a dimly lit single image is appropriate. However, if presented with another image with even weaker lighting, they can easily determine which image is worse. This may explain the notable \textbf{6\%} improvements for junior-level human from single images to image pairs. 

In conclusion, the performance of open-source MLLMs on low-level \textbf{perception} for image pairs is still far from satisfactory, which needs to be enhanced and optimized.

\begin{table*}\small
    \centering
    \renewcommand\arraystretch{1.25}
    \renewcommand\tabcolsep{4.5pt}
    \caption{Main evaluation results on the zero-shot \textbf{Assessment} ability of MLLMs, in comparison with NIQE and CLIP-ViT-Large-14, the visual backbone of most MLLMs. Metrics are \textit{SRCC/PLCC}.}
    \vspace{-5pt}
    \resizebox{\linewidth}{!}{\begin{tabular}{l|cccc|cc|c|c}
    \toprule
    {\textbf{Dataset Type}}  & \multicolumn{4}{c|}{{In-the-wild}} & \multicolumn{2}{c|}{{Generated}} & \multicolumn{1}{c|}{{Artificial}} & \multirow{2}{27pt}{\textit{Average}}\\ \cdashline{1-8}
     \textbf{Model / Dataset}  &{\textit{KONiQ-10k}} & {\textit{SPAQ}} & {\textit{LIVE-FB}} & \textit{LIVE-itw} & {\textit{CGIQA-6K}} & {\textit{AGIQA-3K}} & {\textit{KADID-10K}} & \\ \hline 
    NIQE~\cite{niqe} & 0.316/0.377 & \underline{0.693}/{0.669} & 0.211/0.288 & {0.480}/0.451 & 0.075/0.056 & 0.562/0.517 & 0.374/0.428 & 0.387/0.398\\
    CLIP-ViT-Large-14~\cite{clip} & {0.468}/{0.505} & 0.385/0.389 & 0.218/0.237 & 0.307/0.308 & \underline{0.285}/\underline{0.290} & 0.436/0.458 & 0.376/0.388 & 0.354/0.368\\ \cdashline{1-9}
    InfiMM (\textit{Zephyr-7B})~\cite{InfiMM} & \underline{0.507}/\underline{0.547} & 0.616/0.633 & 0.269/0.299 & \underline{0.548}/\underline{0.580} & 0.229/0.245 & \underline{0.706}/\underline{0.767} & 0.466/0.452 & \underline{0.477}/0.503\\
    Emu2-Chat (\textit{LLaMA-33B})~\cite{emu2} & \textbf{0.664}/\textbf{0.714} & \underline{0.712}/\underline{0.698} & \underline{0.355}/\underline{0.341} & \underline{0.597}/\underline{0.611} & 0.224/0.269 & \textbf{0.759}/{0.751} & \textbf{0.841}/\textbf{0.790} & \textbf{0.593}/\textbf{0.596}\\
    Fuyu-8B (\textit{Persimmon-8B})~\cite{fuyu-8b} & 0.124/0.123 & 0.125/0.179 & 0.164/0.133 & 0.225/0.176 & 0.118/0.116 & 0.368/0.317 & 0.099/0.088 & 0.174/0.161 \\
    BakLLava (\textit{Mistral-7B})~\cite{bakllava} & 0.389/0.390 & 0.406/0.398 & 0.227/0.216 & 0.335/0.337 & 0.179/0.209 & 0.542/0.561 & 0.344/0.361 & 0.346/0.353\\
     mPLUG-Owl2 \textit{(LLaMA-7B)}~\cite{mplug2} & 0.196/0.252 & 0.589/0.614 & 0.217/0.286 & 0.293/0.342 & -0.024/-0.032 & 0.473/0.492  &  0.541/0.546 & 0.326/0.357\\
    LLaVA-v1.5 (\textit{Vicuna-v1.5-7B)}~\cite{improvedllava} & {0.463}/0.459 & 0.443/0.467  & {0.305}/0.321 & 0.344/0.358 & \textbf{0.321}/\textbf{0.333} & {0.672}/{0.738} & 0.417/0.440 & 0.424/0.445\\
    LLaVA-v1.5 (\textit{Vicuna-v1.5-13B)}~\cite{improvedllava} & 0.448/{0.460} & 0.563/0.584  & \underline{0.310}/\underline{0.339} & 0.445/0.481 & \underline{0.285}/\underline{0.297} & 0.664/\underline{0.754} & 0.390/0.400 & 0.444/0.474\\
    InternLM-XComposer-VL \textit{(InternLM)}~\cite{xcomposer}  & \underline{0.564}/\underline{0.615} & \textbf{0.730}/\textbf{0.750} & \textbf{0.360}/\textbf{0.416} & \textbf{0.612}/\textbf{0.676} & 0.243/0.265 & \underline{0.732}/\textbf{0.775} & \underline{0.546}/\underline{0.572} & \underline{0.541}/\underline{0.581}\\
    IDEFICS-Instruct \textit{(LLaMA-7B)}~\cite{idefics} & 0.375/0.400 & 0.474/0.484 & 0.235/0.240 & 0.409/0.428 & 0.244/0.227 & 0.562/0.622 & 0.370/0.373 & 0.381/0.396\\
    Qwen-VL \textit{(QwenLM)}~\cite{Qwen-VL}  & 0.470/0.546 & {0.676}/{0.669} & {0.298}/{0.338} & {0.504}/{0.532} & 0.273/0.284 & 0.617/0.686 & {0.486}/{0.486} & {0.475}/\underline{0.506}\\
    Shikra (\textit{Vicuna-7B)}~\cite{shikra} & 0.314/0.307 & 0.320/0.337 & 0.237/0.241 & 0.322/0.336 & 0.198/0.201 & 0.640/0.661 & 0.324/0.332 & 0.336/0.345\\
    Otter-v1 \textit{(MPT-7B)}~\cite{otter} & 0.406/0.406 & 0.436/0.441 & 0.143/0.142 & -0.008/0.018 & 0.254/0.264 & 0.475/0.481 & \underline{0.557}/\underline{0.577} & 0.323/0.333\\
    Kosmos-2~\cite{kosmos2} & 0.255/0.281 & {0.644}/0.641 & 0.196/0.195 & 0.358/0.368 & 0.210/0.225 & 0.489/0.491 & 0.359/0.365 & 0.359/0.367\\
    InstructBLIP \textit{(Flan-T5-XL)}~\cite{iblip} & 0.334/0.362 & 0.582/0.599 & 0.248/0.267 & 0.113/0.113 & 0.167/0.188 & 0.378/0.400 & 0.211/0.179 & 0.290/0.301\\
    InstructBLIP \textit{(Vicuna-7B)}~\cite{iblip} & 0.359/0.437 & {0.683}/\underline{0.689} & 0.200/0.283 & 0.253/0.367 & 0.263/\underline{0.304} & 0.629/0.663 & 0.337/0.382 & 0.389/0.446\\
    VisualGLM-6B \textit{(GLM-6B)}~\cite{glm}  & 0.247/0.234 & {0.498}/{0.507} & 0.146/0.154 & 0.110/0.116 & 0.209/0.183 & 0.342/0.349 & 0.127/0.131 & 0.240/0.239\\
    mPLUG-Owl \textit{(LLaMA-7B)}~\cite{mplugowl} & 0.409/0.427 & 0.634/{0.644} & 0.241/0.271 & 0.437/{0.487} & 0.148/0.180 & {0.687}/{0.711} & 0.466/{0.486} & 0.432/0.458\\
    LLaMA-Adapter-V2~\cite{llamaadapterv2} & 0.354/0.363 & 0.464/0.506 & 0.275/{0.329} & 0.298/0.360 & 0.257/0.271 & 0.604/0.666 & 0.412/0.425 & 0.381/0.417\\
    LLaVA-v1 \textit{(Vicuna-13B)}~\cite{llava} & 0.462/0.457 & 0.442/0.462 & 0.264/0.280 & 0.404/0.417 & 0.208/0.237 & 0.626/0.684 & 0.349/0.372 & 0.394/0.416\\
    MiniGPT-4 (\textit{Vicuna-13B)}~\cite{minigpt4} & 0.239/0.257 & 0.238/0.253 & 0.170/0.183 & 0.339/0.340 & 0.252/0.246 & 0.572/0.591 & 0.239/0.233 & 0.293/0.300\\
    \bottomrule
    \end{tabular}}
    \vspace{-12pt}
    \label{tab:assessment}
\end{table*}

\begin{figure*}[!t]
    \centering
    \includegraphics[width=\linewidth]{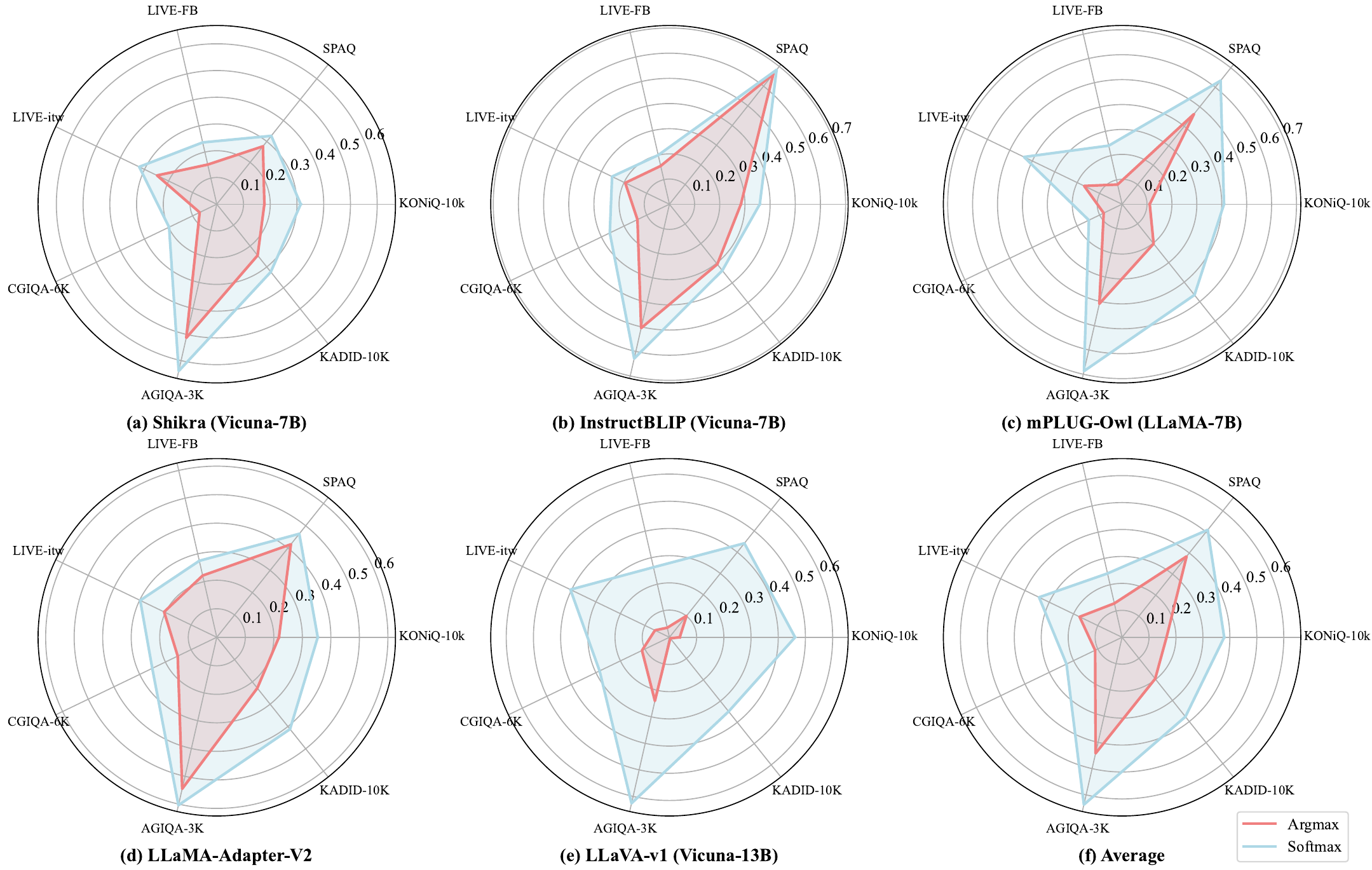}
    \caption{{ Effectiveness radar chart of the proposed {\tt softmax} probability-based strategy against the baseline {\tt argmax} strategy, on multiple MLLMs and different IQA datasets. Performance is $\frac{(SRCC+PLCC)}{2}$.}}
    \label{fig:softmax}
    \vspace{-12pt}
\end{figure*}

\subsection{Findings on \textbf{Description}}

\subsubsection{GPT reliability validation} { We utilize the single-modal GPT to assist in rating the quality of the descriptions, which raises specific concerns about whether GPT is suitable for this task. Consequently, we have specifically conducted experiments regarding the reliability of GPT on the description judgment task. Considering our adoption of a 5-round GPT-assisted evaluation protocol, which averages five GPT-predicted scores, {we invite five human participants to join our user study.} Each participant is presented with the same prompt that GPT encountered and asked to assess the descriptions of selected top-performing MLLMs on a scale of \{0, 1, 2\} for completeness, precision, and relevance. We collect these ratings and average them to establish the ground truth for the descriptions. We sample 100 descriptions in total, comprised of 50 single-image and 50 comparative image-pair descriptions. Subsequently, we compute the SRCC and PLCC values between the ground truth and GPT-predicted scores for comparative analysis. The performance is presented in Table \ref{tab:reliable}. Analysis of this table shows that the scores predicted by GPT strongly correlate with scores rated by humans, as evidenced by the consistently high SRCC, all of which exceed 0.95. We also rank the selected MLLMs by aggregating human ratings and discover that \textbf{the ranking given by humans aligns perfectly with the rankings derived from the GPT-assisted evaluation}. These results indicate that GPT can be recognized as a reliable tool for evaluating descriptions.}

\subsubsection{Description for single images} For the \textbf{description} ability exhibited in Table \ref{tab:description}, InternLM-XComposer-VL reaches the best proficiency, especially in terms of the relevance dimension. Nevertheless, in the perspective of the completeness and precision of the descriptions, even the best of all MLLMs cannot obtain an excellent score; on the contrary, almost all MLLMs reach an acceptable standard (0.8/2.0). In general, all MLLMs at present are only with relatively limited and primary ability to provide low-level visual descriptions.

\subsubsection{Description for image pairs} We also include the mPLUG-Owl2 (\textit{Q-Instruct}) for \textbf{description-pair} ability benchmark. As shown in Table \ref{tab:description_pair}, similarly, all MLLMs perform better in the aspect of relevance than completeness and precision. Furthermore, fine-tuned with the single image low-level visual dataset \textbf{Q-Instruct} \cite{q-instruct},  mPLUG-Owl2 (\textit{Q-Instruct}) achieves the best performance on completeness and relevance but gets the lowest score on precision. This indicates the knowledge infusion from single images can effectively enhance an MLLM to focus on corresponding low-level dimensions for targeted responses, but it does not improve the accuracy of the content, meaning it cannot enhance the core analytical ability of image pairs.

\subsection{Findings on \textbf{Assessment}}

\begin{figure*}
    \centering
    \includegraphics[width=\linewidth]{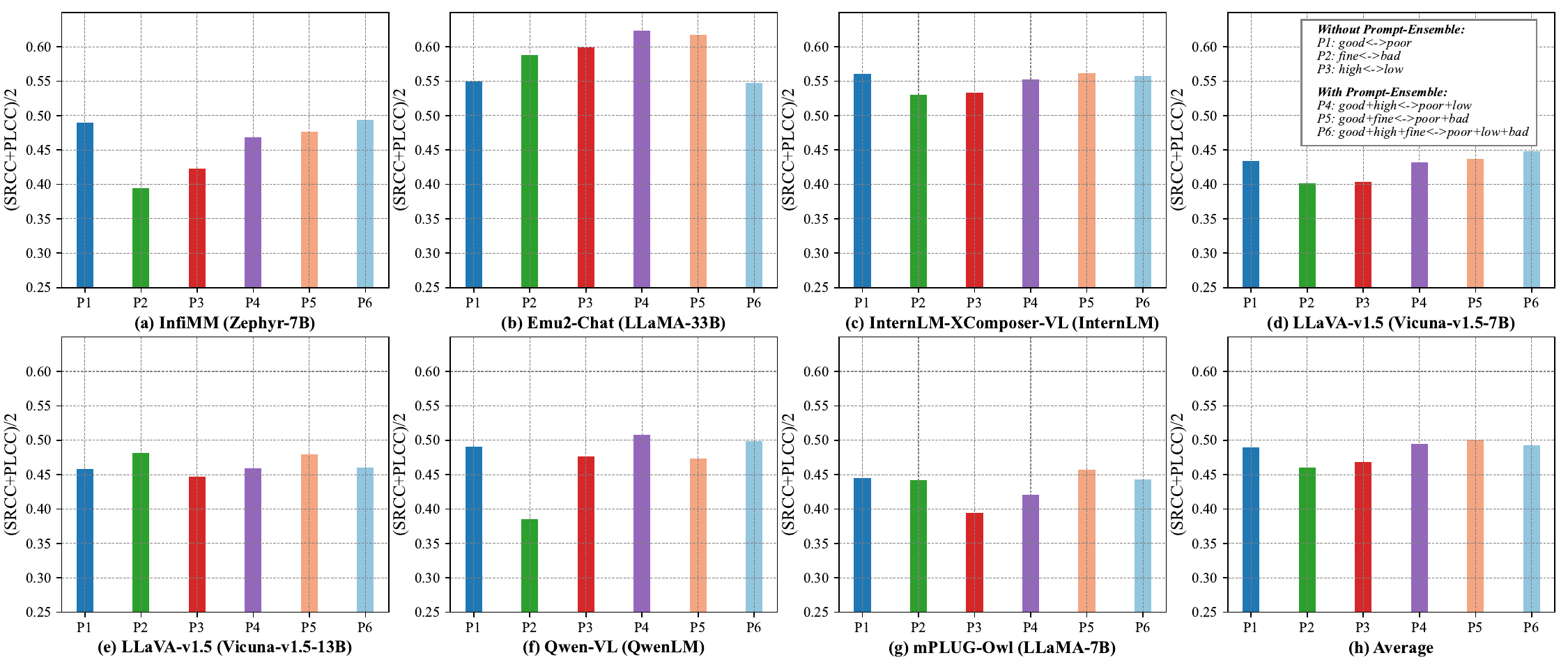}
    \caption{{ Evaluation results on the \textit{prompt ensemble} strategy for the \textbf{Assessment} ability on MLLMs with top-7 results in the default A3 leaderboard of the \textbf{Q-Bench$^+$}. The performance is recorded as the average $\frac{(SRCC+PLCC)}{2}$ value across the 7 quality assessment datasets. After \textit{ensemble}, the rankings among them are not changed.}}
    \label{fig:ensemble}
\end{figure*}

\subsubsection{MLLM Performance}
To measure the \textbf{assessment} ability, we evaluate the performance of 20 open-source MLLMs on 7 IQA datasets that are with at least \textbf{1,000} images and \textbf{15} human ratings per image~\cite{itu}. The experimental results are illustrated in Table \ref{tab:assessment}. a) Primarily, we notice that the majority of MLLMs are notably better than NIQE on \textbf{non-natural} circumstances (CGI, AIGC, artificial distortions), showing their potential towards general-purpose evaluators on a broader range of low-level appearances. b) We also notice that without explicit alignment with human opinions during training, the most excellent MLLM, Emu2-Chat (\textit{which is based on the heaviest LLM, LLaMA-33B}), can already outperform CLIP-ViT-Large-14 by a large margin (\textbf{25\%}). {These results have demonstrated that, though most MLLMs are still based on CLIP as visual encoders, their high capacity in the strong language decoder can do help them perform much better on visual quality assessment even without any explicit training.}

\subsubsection{Superiority of {\tt softmax}}
{ In this section, we quantitatively evaluate the correlation with human perception on a simple {\tt argmax} strategy between \textit{good$\leftrightarrow$bad} and our proposed {\tt softmax} strategy. In Fig.~\ref{fig:softmax}, we select 5 MLLMs of different architectures and confirm that for all IQA datasets, the more measurable {\tt softmax} strategy predicts better than the {\tt argmax} strategy, which degenerates into only two scores, 0 and 1. Though the result is generally expected, the experiments validate that MLLMs have quantitative \textbf{assessment} ability hidden behind their word outputs, and prove the effectiveness of our softmax-based IQA strategy.}

\subsubsection{Prompt Ensemble Effectiveness}
{ As shown in Fig.~\ref{fig:ensemble}, the \textit{prompt ensemble} strategy (as proposed in Eq.~\ref{eq:ensemble}) on top-7 MLLMs (\textit{i.e.} Emu2-Chat, InternLM-XComposer-VL, Qwen-VL, InfiMM, LLaVA-v1.5 (\textit{13B}), mPLUG-Owl, and LLaVA-v1.5 (\textit{7B})) can lead to up to 5\% accuracy improvement (\textit{in average 1.7\%}). We believe it is a useful boost technique to improve the performance of MLLMs on the IQA task.
Nevertheless, we also notice that different MLLMs perform best with different specific prompt combos. For example, the \textit{good+fine}$\leftrightarrow$\textit{poor+bad} performs best on InternLM-XComposer-VL, but comes with reduced accuracy on QWen-VL compared with only \textit{good$\leftrightarrow$poor}. While \textit{good}$\leftrightarrow$\textit{poor} is proved \textit{overall best single word pair} (except Emu2-Chat and LLaVA-v1.5 (\textit{13B})) for the evaluation and shows stable results across MLLMs, we decide to keep the current strategy (using the \textit{good}$\leftrightarrow$\textit{poor} combo) in \textbf{Q-Bench$^+$}.}

\section{Conclusion}
In this research, we introduce \textbf{Q-Bench$^+$}, a benchmark designed to evaluate the advancements of MLLMs in low-level visual abilities. We evaluate the MLLMs from three aspects: \textbf{perception} of low-level visual attributes, \textbf{description} of low-level visual content, and \textbf{assessment} of image quality. Additionally, acknowledging the importance of discerning differences and similarities in image pairs, our benchmark encompasses both single images and image pairs in the \textbf{perception} and \textbf{description} tasks. To evaluate these abilities, we have compiled two multi-modal benchmark datasets focused on low-level vision, and introduced a unified softmax-based method for quantitative image quality assessment (IQA) in MLLMs. Our findings demonstrate that some advanced MLLMs exhibit commendable low-level visual abilities even without specialized low-level training. However, there's still a significant journey ahead before MLLMs can become fully reliable assistants in general low-level visual tasks. We hope that the insights gained from \textbf{Q-Bench$^+$} will spur further development in MLLMs, particularly in improving their perception and understanding of low-level visual elements.

\bibliographystyle{IEEEtran}
\bibliography{ref}

\begin{thebibliography}{10}
\providecommand{\url}[1]{#1}
\csname url@samestyle\endcsname
\providecommand{\newblock}{\relax}
\providecommand{\bibinfo}[2]{#2}
\providecommand{\BIBentrySTDinterwordspacing}{\spaceskip=0pt\relax}
\providecommand{\BIBentryALTinterwordstretchfactor}{4}
\providecommand{\BIBentryALTinterwordspacing}{\spaceskip=\fontdimen2\font plus
\BIBentryALTinterwordstretchfactor\fontdimen3\font minus \fontdimen4\font\relax}
\providecommand{\BIBforeignlanguage}[2]{{%
\expandafter\ifx\csname l@#1\endcsname\relax
\typeout{** WARNING: IEEEtran.bst: No hyphenation pattern has been}%
\typeout{** loaded for the language `#1'. Using the pattern for}%
\typeout{** the default language instead.}%
\else
\language=\csname l@#1\endcsname
\fi
#2}}
\providecommand{\BIBdecl}{\relax}
\BIBdecl

\bibitem{llama}
H.~Touvron, T.~Lavril, G.~Izacard, X.~Martinet, M.-A. Lachaux, T.~Lacroix, B.~Rozière, N.~Goyal, E.~Hambro, F.~Azhar, A.~Rodriguez, A.~Joulin, E.~Grave, and G.~Lample, ``Llama: Open and efficient foundation language models,'' 2023.

\bibitem{mpt}
\BIBentryALTinterwordspacing
M.~N. Team. (2023) Introducing mpt-7b: A new standard for open-source, commercially usable llms. Accessed: 2023-05-05. [Online]. Available: \url{www.mosaicml.com/blog/mpt-7b}
\BIBentrySTDinterwordspacing

\bibitem{llava}
H.~Liu, C.~Li, Q.~Wu, and Y.~J. Lee, ``Visual instruction tuning,'' 2023.

\bibitem{minigpt4}
D.~Zhu, J.~Chen, X.~Shen, X.~Li, and M.~Elhoseiny, ``Minigpt-4: Enhancing vision-language understanding with advanced large language models,'' \emph{arXiv preprint arXiv:2304.10592}, 2023.

\bibitem{iblip}
W.~Dai, J.~Li, D.~Li, A.~M.~H. Tiong, J.~Zhao, W.~Wang, B.~Li, P.~Fung, and S.~Hoi, ``Instructblip: Towards general-purpose vision-language models with instruction tuning,'' 2023.

\bibitem{otter}
B.~Li, Y.~Zhang, L.~Chen, J.~Wang, J.~Yang, and Z.~Liu, ``Otter: A multi-modal model with in-context instruction tuning,'' \emph{arXiv preprint arXiv:2305.03726}, 2023.

\bibitem{cococaps}
X.~Chen, H.~Fang, T.-Y. Lin, R.~Vedantam, S.~Gupta, P.~Dollar, and C.~L. Zitnick, ``Microsoft coco captions: Data collection and evaluation server,'' 2015.

\bibitem{cocovqa}
S.~Antol, A.~Agrawal, J.~Lu, M.~Mitchell, D.~Batra, C.~L. Zitnick, and D.~Parikh, ``{VQA}: {V}isual {Q}uestion {A}nswering,'' in \emph{ICCV}, 2015.

\bibitem{kosmos2}
Z.~Peng, W.~Wang, L.~Dong, Y.~Hao, S.~Huang, S.~Ma, and F.~Wei, ``Kosmos-2: Grounding multimodal large language models to the world,'' \emph{ArXiv}, vol. abs/2306, 2023.

\bibitem{lai2023lisa}
X.~Lai, Z.~Tian, Y.~Chen, Y.~Li, Y.~Yuan, S.~Liu, and J.~Jia, ``Lisa: Reasoning segmentation via large language model,'' \emph{arXiv preprint arXiv:2308.00692}, 2023.

\bibitem{koniq}
V.~Hosu, H.~Lin, T.~Sziranyi, and D.~Saupe, ``Koniq-10k: An ecologically valid database for deep learning of blind image quality assessment,'' \emph{IEEE TIP}, vol.~29, pp. 4041--4056, 2020.

\bibitem{spaq}
Y.~Fang, H.~Zhu, Y.~Zeng, K.~Ma, and Z.~Wang, ``Perceptual quality assessment of smartphone photography,'' in \emph{CVPR}, 2020.

\bibitem{koniqplusplus}
S.~Su, V.~Hosu, H.~Lin, Y.~Zhang, and D.~Saupe, ``Koniq++ : Boosting no-reference image quality assessment in the wild by jointly predicting image quality and defects,'' in \emph{The British Machine Vision Conference (BMVC)}, 2021, pp. 1--12.

\bibitem{wu2023explainable}
H.~Wu, E.~Zhang, L.~Liao, C.~Chen, J.~Hou, A.~Wang, W.~Sun, Q.~Yan, and W.~Lin, ``Towards explainable video quality assessment: A database and a language-prompted approach,'' in \emph{ACM MM}, 2023.

\bibitem{aadb}
S.~Kong, X.~Shen, Z.~Lin, R.~Mech, and C.~Fowlkes, ``Photo aesthetics ranking network with attributes and content adaptation,'' in \emph{ECCV}, 2016.

\bibitem{avaiaa}
N.~Murray, L.~Marchesotti, and F.~Perronnin, ``Ava: A large-scale database for aesthetic visual analysis,'' in \emph{CVPR}, 2012, pp. 2408--2415.

\bibitem{zhang2023subjective}
Z.~Zhang, W.~Sun, T.~Wang, W.~Lu, Q.~Zhou, Q.~Wang, X.~Min, G.~Zhai \emph{et~al.}, ``Subjective and objective quality assessment for in-the-wild computer graphics images,'' \emph{ACM TOMM}, 2023.

\bibitem{agiqa3k}
C.~Li, Z.~Zhang, H.~Wu, W.~Sun, X.~Min, X.~Liu, G.~Zhai, and W.~Lin, ``Agiqa-3k: An open database for ai-generated image quality assessment,'' 2023.

\bibitem{imagereward}
J.~Xu, X.~Liu, Y.~Wu, Y.~Tong, Q.~Li, M.~Ding, J.~Tang, and Y.~Dong, ``Imagereward: Learning and evaluating human preferences for text-to-image generation,'' 2023.

\bibitem{wu2023dover}
H.~Wu, E.~Zhang, L.~Liao, C.~Chen, J.~Hou, A.~Wang, W.~Sun, Q.~Yan, and W.~Lin, ``Exploring video quality assessment on user generated contents from aesthetic and technical perspectives,'' in \emph{ICCV}, 2023.

\bibitem{irpotential}
C.~Zhang, S.~Su, Y.~Zhu, Q.~Yan, J.~Sun, and Y.~Zhang, ``Exploring and evaluating image restoration potential in dynamic scenes,'' in \emph{CVPR}, 2022, pp. 2057--2066.

\bibitem{lpips}
R.~Zhang, P.~Isola, A.~A. Efros, E.~Shechtman, and O.~Wang, ``The unreasonable effectiveness of deep features as a perceptual metric,'' in \emph{CVPR}, 2018, pp. 586--595.

\bibitem{pieapp}
E.~Prashnani, H.~Cai, Y.~Mostofi, and P.~Sen, ``Pieapp: Perceptual image-error assessment through pairwise preference,'' in \emph{CVPR}, June 2018.

\bibitem{pipal}
J.~Gu, H.~Cai, H.~Chen, X.~Ye, J.~Ren, and C.~Dong, ``Pipal: a large-scale image quality assessment dataset for perceptual image restoration,'' 2020.

\bibitem{emu2}
Q.~Sun, Y.~Cui, X.~Zhang, F.~Zhang, Q.~Yu, Z.~Luo, Y.~Wang, Y.~Rao, J.~Liu, T.~Huang \emph{et~al.}, ``Generative multimodal models are in-context learners,'' \emph{arXiv preprint arXiv:2312.13286}, 2023.

\bibitem{bakllava}
\BIBentryALTinterwordspacing
SkunkworksAI, ``Bakllava,'' 2024. [Online]. Available: \url{https://github.com/SkunkworksAI/BakLLaVA}
\BIBentrySTDinterwordspacing

\bibitem{geminipro}
\BIBentryALTinterwordspacing
Google, ``Gemini pro,'' 2023. [Online]. Available: \url{https://deepmind.google/technologies/gemini}
\BIBentrySTDinterwordspacing

\bibitem{openai2023gpt4}
OpenAI, ``Gpt-4 technical report,'' 2023.

\bibitem{mmbench}
Y.~Liu, H.~Duan, Y.~Zhang, B.~Li, S.~Zhang, W.~Zhao, Y.~Yuan, J.~Wang, C.~He, Z.~Liu, K.~Chen, and D.~Lin, ``Mmbench: Is your multi-modal model an all-around player?'' 2023.

\bibitem{emabench}
J.~Lu, J.~Rao, K.~Chen, X.~Guo, Y.~Zhang, B.~Sun, C.~Yang, and J.~Yang, ``Evaluation and mitigation of agnosia in multimodal large language models,'' 2023.

\bibitem{atqa}
T.~Guha, V.~Hosu, D.~Saupe, B.~Goldl\"{u}cke, N.~Kumar, W.~Lin, V.~Martinez, K.~Somandepalli, S.~Narayanan, W.-H. Cheng, K.~McLaughlin, H.~Adam, J.~See, and L.-K. Wong, ``Atqam/mast'20: Joint workshop on aesthetic and technical quality assessment of multimedia and media analytics for societal trends,'' in \emph{ACM MM}, 2020, p. 4758–4760.

\bibitem{sfa}
D.~Li, T.~Jiang, W.~Lin, and M.~Jiang, ``Which has better visual quality: The clear blue sky or a blurry animal?'' \emph{IEEE TMM}, vol.~21, no.~5, pp. 1221--1234, 2019.

\bibitem{kadid}
H.~Lin, V.~Hosu, and D.~Saupe, ``Kadid-10k: A large-scale artificially distorted iqa database,'' in \emph{QoMEX}, 2019, pp. 1--3.

\bibitem{livechallenge}
D.~Ghadiyaram and A.~C. Bovik, ``Massive online crowdsourced study of subjective and objective picture quality,'' \emph{IEEE TIP}, vol.~25, no.~1, pp. 372--387, 2015.

\bibitem{wu2024qbench}
H.~Wu, Z.~Zhang, E.~Zhang, C.~Chen, L.~Liao, A.~Wang, C.~Li, W.~Sun, Q.~Yan, G.~Zhai, and W.~Lin, ``Q-bench: A benchmark for general-purpose foundation models on low-level vision,'' in \emph{ICLR}, 2024.

\bibitem{imagecorruptions}
C.~Michaelis, B.~Mitzkus, R.~Geirhos, E.~Rusak, O.~Bringmann, A.~S. Ecker, M.~Bethge, and W.~Brendel, ``Benchmarking robustness in object detection: Autonomous driving when winter is coming,'' \emph{arXiv preprint arXiv:1907.07484}, 2019.

\bibitem{paq2piq}
Z.~Ying, H.~Niu, P.~Gupta, D.~Mahajan, D.~Ghadiyaram, and A.~Bovik, ``From patches to pictures (paq-2-piq): Mapping the perceptual space of picture quality,'' in \emph{CVPR}, 2020.

\bibitem{clive}
D.~Ghadiyaram and A.~C. Bovik, ``Massive online crowdsourced study of subjective and objective picture quality,'' \emph{IEEE}, vol.~25, no.~1, pp. 372--387, 2016.

\bibitem{livemultipledistortions}
D.~Jayaraman, A.~Mittal, A.~K. Moorthy, and A.~C. Bovik, ``Objective quality assessment of multiply distorted images,'' in \emph{ASILOMAR}, 2012, pp. 1693--1697.

\bibitem{seedbench}
B.~Li, R.~Wang, G.~Wang, Y.~Ge, Y.~Ge, and Y.~Shan, ``Seed-bench: Benchmarking multimodal llms with generative comprehension,'' 2023.

\bibitem{okvqa}
K.~Marino, M.~Rastegari, A.~Farhadi, and R.~Mottaghi, ``Ok-vqa: A visual question answering benchmark requiring external knowledge,'' in \emph{Conference on Computer Vision and Pattern Recognition (CVPR)}, 2019.

\bibitem{clipiaa}
J.~Hou, W.~Lin, Y.~Fang, H.~Wu, C.~Chen, L.~Liao, and W.~Liu, ``Towards transparent deep image aesthetics assessment with tag-based content descriptors,'' \emph{IEEE TIP}, 2023.

\bibitem{nima}
H.~Talebi and P.~Milanfar, ``Nima: Neural image assessment,'' \emph{IEEE TIP}, 2018.

\bibitem{rfugc}
Y.~Wang, J.~Ke, H.~Talebi, J.~G. Yim, N.~Birkbeck, B.~Adsumilli, P.~Milanfar, and F.~Yang, ``Rich features for perceptual quality assessment of ugc videos,'' in \emph{CVPR}, June 2021, pp. 13\,435--13\,444.

\bibitem{wu2022fastervqa}
H.~Wu, C.~Chen, L.~Liao, J.~Hou, W.~Sun, Q.~Yan, J.~Gu, and W.~Lin, ``Neighbourhood representative sampling for efficient end-to-end video quality assessment,'' 2023.

\bibitem{qalign}
H.~Wu, Z.~Zhang, W.~Zhang, C.~Chen, L.~Liao, C.~Li, Y.~Gao, A.~Wang, E.~Zhang, W.~Sun \emph{et~al.}, ``Q-align: Teaching lmms for visual scoring via discrete text-defined levels,'' \emph{arXiv preprint arXiv:2312.17090}, 2023.

\bibitem{fastvqa}
H.~Wu, C.~Chen, J.~Hou, L.~Liao, A.~Wang, W.~Sun, Q.~Yan, and W.~Lin, ``Fast-vqa: Efficient end-to-end video quality assessment with fragment sampling,'' in \emph{ECCV}, 2022.

\bibitem{pvq}
Z.~Ying, M.~Mandal, D.~Ghadiyaram, and A.~Bovik, ``Patch-vq: 'patching up' the video quality problem,'' in \emph{CVPR}, 2021.

\bibitem{hore2010image}
A.~Hore and D.~Ziou, ``Image quality metrics: Psnr vs. ssim,'' in \emph{International Conference on Pattern Recognition}.\hskip 1em plus 0.5em minus 0.4em\relax IEEE, 2010, pp. 2366--2369.

\bibitem{zhang2021no}
Z.~Zhang, W.~Sun, X.~Min, W.~Zhu, T.~Wang, W.~Lu, and G.~Zhai, ``A no-reference evaluation metric for low-light image enhancement,'' in \emph{IEEE International Conference on Multimedia and Expo}.\hskip 1em plus 0.5em minus 0.4em\relax IEEE, 2021, pp. 1--6.

\bibitem{zhang2022no}
Z.~Zhang, W.~Sun, X.~Min, W.~Zhu, T.~Wang, and G.~Zhai, ``A no-reference deep learning quality assessment method for super-resolution images based on frequency maps,'' in \emph{IEEE International Symposium on Circuits and Systems}, 2022, pp. 3170--3174.

\bibitem{mplugowl}
Q.~Ye, H.~Xu, G.~Xu, J.~Ye, M.~Yan, Y.~Zhou, J.~Wang, A.~Hu, P.~Shi, Y.~Shi, C.~Jiang, C.~Li, Y.~Xu, H.~Chen, J.~Tian, Q.~Qi, J.~Zhang, and F.~Huang, ``mplug-owl: Modularization empowers large language models with multimodality,'' 2023.

\bibitem{flickrcaps}
P.~Young, A.~Lai, M.~Hodosh, and J.~Hockenmaier, ``From image descriptions to visual denotations: New similarity metrics for semantic inference over event descriptions,'' \emph{Transactions of the Association for Computational Linguistics}, vol.~2, pp. 67--78, 2014.

\bibitem{nocaps}
H.~Agrawal, K.~Desai, Y.~Wang, X.~Chen, R.~Jain, M.~Johnson, D.~Batra, D.~Parikh, S.~Lee, and P.~Anderson, ``nocaps: novel object captioning at scale,'' in \emph{ICCV}, 2019.

\bibitem{vicuna}
L.~Zheng, W.-L. Chiang, Y.~Sheng, S.~Zhuang, Z.~Wu, Y.~Zhuang, Z.~Lin, Z.~Li, D.~Li, E.~P. Xing, H.~Zhang, J.~E. Gonzalez, and I.~Stoica, ``Judging llm-as-a-judge with mt-bench and chatbot arena,'' 2023.

\bibitem{InfiMM}
\BIBentryALTinterwordspacing
I.~Team, ``Infimm: Advancing multimodal understanding from flamingo's legacy through diverse llm integration,'' 2024. [Online]. Available: \url{https://huggingface.co/Infi-MM/}
\BIBentrySTDinterwordspacing

\bibitem{fuyu-8b}
\BIBentryALTinterwordspacing
R.~Bavishi, E.~Elsen, C.~Hawthorne, M.~Nye, A.~Odena, A.~Somani, and S.~Ta\c{s}\i{}rlar, ``Introducing our multimodal models,'' 2023. [Online]. Available: \url{https://www.adept.ai/blog/fuyu-8b}
\BIBentrySTDinterwordspacing

\bibitem{sphinx}
Z.~Lin, C.~Liu, R.~Zhang, P.~Gao, L.~Qiu, H.~Xiao, H.~Qiu, C.~Lin, W.~Shao, K.~Chen \emph{et~al.}, ``Sphinx: The joint mixing of weights, tasks, and visual embeddings for multi-modal large language models,'' \emph{arXiv preprint arXiv:2311.07575}, 2023.

\bibitem{mplug2}
Q.~Ye, H.~Xu, J.~Ye, M.~Yan, H.~Liu, Q.~Qian, J.~Zhang, F.~Huang, and J.~Zhou, ``mplug-owl2: Revolutionizing multi-modal large language model with modality collaboration,'' \emph{arXiv preprint arXiv:2311.04257}, 2023.

\bibitem{improvedllava}
H.~Liu, C.~Li, Y.~Li, and Y.~J. Lee, ``Improved baselines with visual instruction tuning,'' 2023.

\bibitem{xcomposer}
P.~Zhang, X.~Dong, B.~Wang, Y.~Cao, C.~Xu, L.~Ouyang, Z.~Zhao, S.~Ding, S.~Zhang, H.~Duan, W.~Zhang, H.~Yan, X.~Zhang, W.~Li, J.~Li, K.~Chen, C.~He, X.~Zhang, Y.~Qiao, D.~Lin, and J.~Wang, ``Internlm-xcomposer: A vision-language large model for advanced text-image comprehension and composition,'' 2023.

\bibitem{idefics}
\BIBentryALTinterwordspacing
Huggingface, ``Introducing idefics: An open reproduction of state-of-the-art visual language model,'' 2023. [Online]. Available: \url{https://huggingface.co/blog/idefics}
\BIBentrySTDinterwordspacing

\bibitem{Qwen-VL}
J.~Bai, S.~Bai, S.~Yang, S.~Wang, S.~Tan, P.~Wang, J.~Lin, C.~Zhou, and J.~Zhou, ``Qwen-vl: A versatile vision-language model for understanding, localization, text reading, and beyond,'' \emph{arXiv preprint arXiv:2308.12966}, 2023.

\bibitem{shikra}
K.~Chen, Z.~Zhang, W.~Zeng, R.~Zhang, F.~Zhu, and R.~Zhao, ``Shikra: Unleashing multimodal llm's referential dialogue magic,'' \emph{arXiv preprint arXiv:2306.15195}, 2023.

\bibitem{glm}
Z.~Du, Y.~Qian, X.~Liu, M.~Ding, J.~Qiu, Z.~Yang, and J.~Tang, ``Glm: General language model pretraining with autoregressive blank infilling,'' in \emph{Proceedings of the 60th Annual Meeting of the Association for Computational Linguistics (Volume 1: Long Papers)}, 2022, pp. 320--335.

\bibitem{llamaadapterv2}
P.~Gao, J.~Han, R.~Zhang, Z.~Lin, S.~Geng, A.~Zhou, W.~Zhang, P.~Lu, C.~He, X.~Yue, H.~Li, and Y.~Qiao, ``Llama-adapter v2: Parameter-efficient visual instruction model,'' \emph{arXiv preprint arXiv:2304.15010}, 2023.

\bibitem{q-instruct}
H.~Wu, Z.~Zhang, E.~Zhang, C.~Chen, L.~Liao, A.~Wang, K.~Xu, C.~Li, J.~Hou, G.~Zhai \emph{et~al.}, ``Q-instruct: Improving low-level visual abilities for multi-modality foundation models,'' \emph{arXiv preprint arXiv:2311.06783}, 2023.

\bibitem{wu2022fastvqa}
H.~Wu, C.~Chen, J.~Hou, L.~Liao, A.~Wang, W.~Sun, Q.~Yan, and W.~Lin, ``Fast-vqa: Efficient end-to-end video quality assessment with fragment sampling,'' in \emph{ECCV}, 2022.

\bibitem{niqe}
A.~Mittal, R.~Soundararajan, and A.~C. Bovik, ``Making a “completely blind” image quality analyzer,'' \emph{IEEE Signal Processing Letters}, vol.~20, no.~3, pp. 209--212, 2013.

\bibitem{clip}
A.~Radford, J.~W. Kim, C.~Hallacy, A.~Ramesh, G.~Goh, S.~Agarwal, G.~Sastry, A.~Askell, P.~Mishkin, J.~Clark, G.~Krueger, and I.~Sutskever, ``Learning transferable visual models from natural language supervision,'' 2021.

\bibitem{itu}
``Recommendation 500-10: Methodology for the subjective assessment of the quality of television pictures,'' ITU-R Rec. BT.500, 2000.

\bibitem{clipiqa}
J.~Wang, K.~C.~K. Chan, and C.~C. Loy, ``Exploring clip for assessing the look and feel of images,'' 2022.

\bibitem{co_instruct}
H.~Wu, H.~Zhu, Z.~Zhang, E.~Zhang, C.~Chen, L.~Liao, C.~Li, A.~Wang, W.~Sun, Q.~Yan \emph{et~al.}, ``Towards open-ended visual quality comparison,'' \emph{arXiv preprint arXiv:2402.16641}, 2024.

\end{thebibliography}

\begin{IEEEbiography}[{\includegraphics[width=1in,height=1.25in,clip,keepaspectratio]{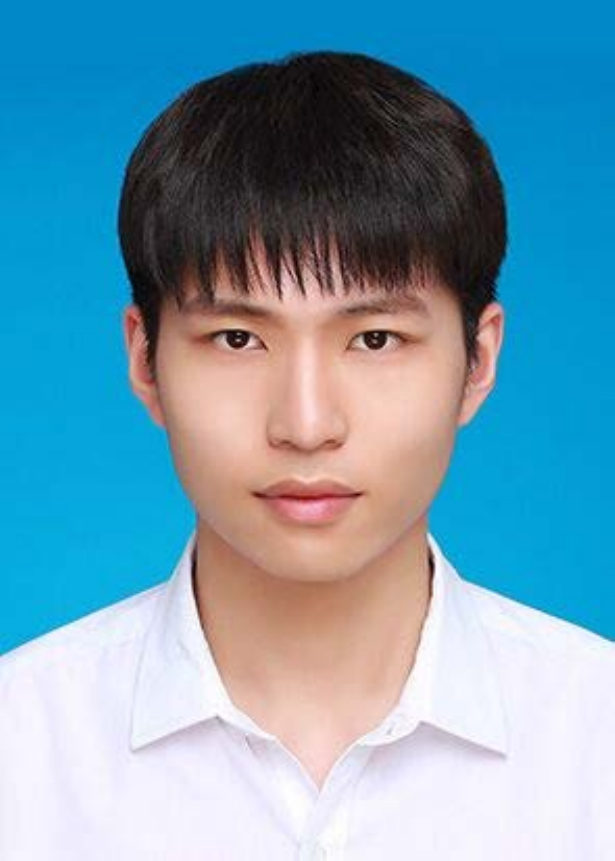}}]{Zicheng Zhang}

Zicheng Zhang received his B.E. degree from Shanghai Jiaotong University, Shanghai, China, in 2020 and he is currently pursuing PhD degree at Shanghai Jiao Tong University. His research interests include quality assessment, low-level vision, and large multi-modal models.

\end{IEEEbiography}

\begin{IEEEbiography}[{\includegraphics[width=1in,height=1.25in,clip,keepaspectratio]{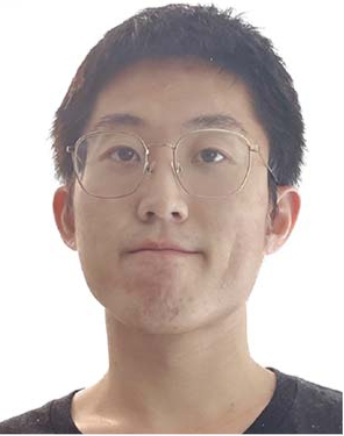}}]{Haoning Wu}

Haoning Wu received the BS degree from the School of Electronic Engineering and Computer Science, Peking University, Beijing, China, in 2021. He is currently working toward the PhD degree with S-Lab, School of Computer Science and Engineering. Nanyang Technological University, Singapore, supervised by Prof. Weisi Lin (Fellow, lEEE). His research interests mainly include video quality assessment, including improving its robustness, efficiency, and interpretability and connecting it with related tasks.

\end{IEEEbiography}

\begin{IEEEbiography}[{\includegraphics[width=1in,height=1.25in,clip,keepaspectratio]{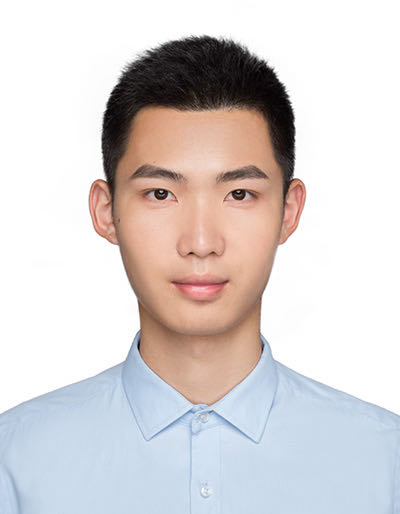}}]{Erli Zhang}

Erli Zhang received his B.E. degree from Nanyang Technological University, Singapore, in 2024. He is about to begin his Ph.D. at the National University of Singapore. His research interests include large multimodal models and AI for healthcare.

\end{IEEEbiography}

\begin{IEEEbiography}[{\includegraphics[width=1in,height=1.25in,clip,keepaspectratio]{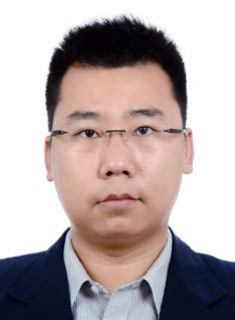}}]{Guangtao Zhai}

Guangtao Zhai (SM’19) received the B.E. and M.E. degrees from Shandong University, Shandong, China, in 2001 and 2004, respectively, and the Ph.D. degree from Shanghai Jiao Tong University, Shanghai, China, in 2009, where he is currently a Research Professor with the Institute of Image Communication and Information Processing. From 2008 to 2009, he was a Visiting Student with the Department of Electrical and Computer Engineering, McMaster University, Hamilton, ON, Canada, where he was a Post-Doctoral Fellow from 2010 to 2012. From 2012 to 2013, he was a Humboldt Research Fellow with the Institute of Multimedia Communication and Signal Processing, Friedrich Alexander University of Erlangen-Nuremberg, Germany. He received the Award of National Excellent Ph.D. Thesis from the Ministry of Education of China in 2012. His research interests include multimedia signal processing and perceptual signal processing.

\end{IEEEbiography}

\begin{IEEEbiography}[{\includegraphics[width=1in,height=1.25in,clip,keepaspectratio]{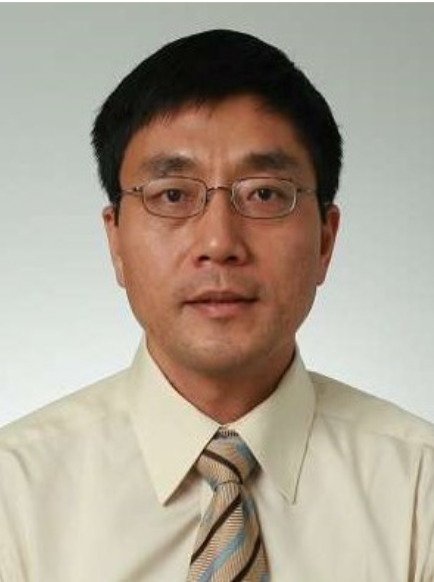}}]{Weisi Lin}

Weisi Lin (Fellow, IEEE) received the bachelor’s degree in electronics and the master’s degree in digital signal processing from Sun Yat-Sen University, Guangzhou, China, and the Ph.D. degree in computer vision from King’s College London, U.K. He is currently a Professor with the School of Computer Science and Engineering, Nanyang Technological University, Singapore. His research interests include image processing, perceptual modeling, video compression, multimedia communication, and computer vision. He is a fellow of the IET, an Honorary Fellow of the Singapore Institute of Engineering Technologists, and a Chartered Engineer in U.K. He was awarded as a Distinguished Lecturer of the IEEE Circuits and Systems Society from 2016 to 2017. He served as a Lead Guest Editor for a Special Issue on Perceptual Signal Processing of the IEEE JSTSP in 2012. He has also served or serves as an Associate Editor for IEEE TIP, IEEE TCSVT, IEEE TMM, IEEE TNNLS, IEEE SPL, and Journal of Visual Communication and Image Representation. He was the Chair of the IEEE MMTC Special Interest Group on Quality of Experience.

\end{IEEEbiography}
\newpage

\clearpage

\section{Appendix}

\subsection{Subjective Experiment}

\begin{figure}[!t]
    \centering
    \subfigure[Interface for the \textbf{LLVisionQA$^+$} dataset (\textbf{Perception})]{\begin{minipage}[t]{\linewidth}
                \centering
                \includegraphics[width = 1\linewidth]{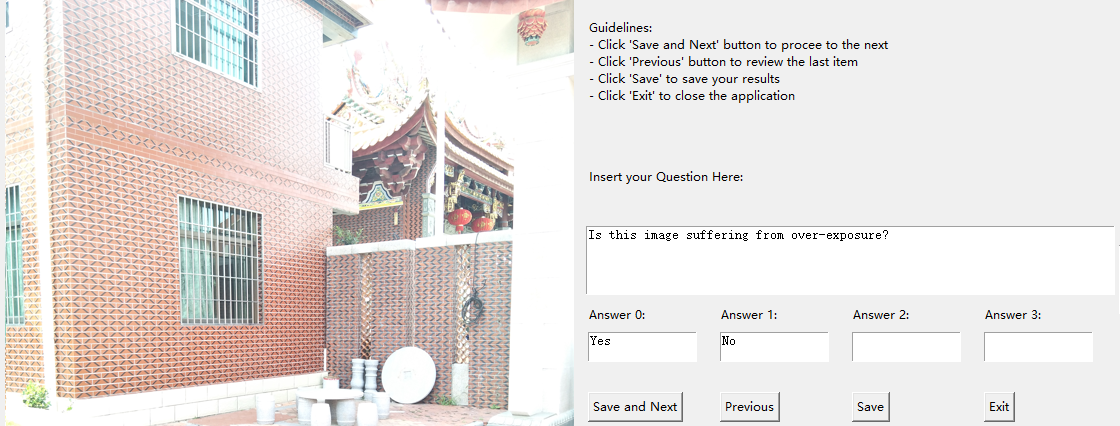}
                \end{minipage}}
    \subfigure[Interface for the \textbf{LLDescribe$^+$} dataset (\textbf{Description})]{\begin{minipage}[t]{\linewidth}
                \centering
                \includegraphics[width = 1\linewidth]{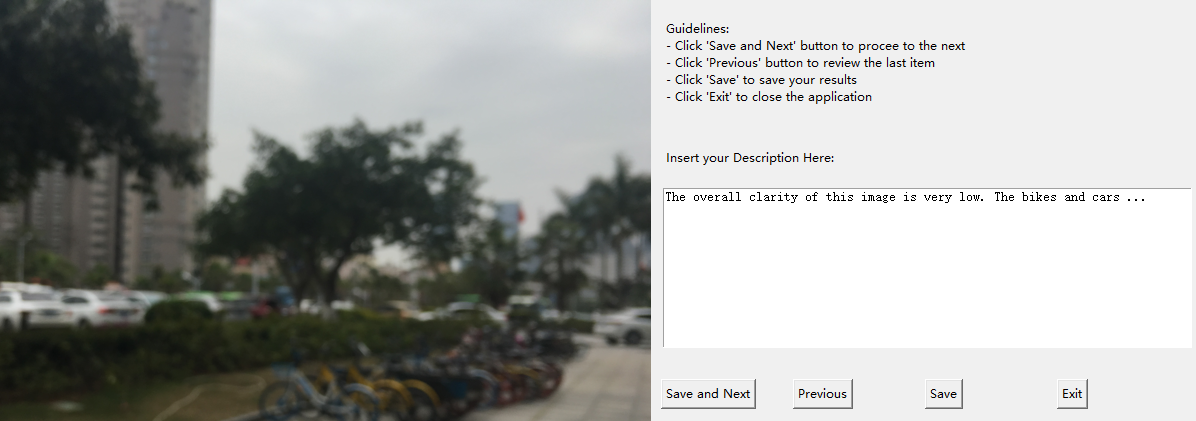}
                \end{minipage}}
    \caption{The illustration of the annotation interfaces for the \textbf{LLVisionQA$^+$} dataset (\textit{questions, answers}) on \textbf{Peception} ability, and the \textbf{LLDescribe$^+$} dataset (\textit{text description}) on \textbf{Description} ability. For image pairs, we simply concatenate the two images into one image for presentation. }
    \label{fig:gui}
    \vspace{-10pt}
\end{figure}

A total of twenty experts, each with professional skills and extensive experience in photography, are invited to participate in the subjective labeling experiment of \textbf{Q-Bench$^+$}. The subjective experiment takes place in a laboratory environment with standard indoor lighting. A Dell-4K monitor, which supports a resolution of $3840\times2160$, is used for displaying the interfaces. The screenshots of interfaces can be referred to in Fig. \ref{fig:gui}. Each expert annotates up to 30 images a day to avoid fatigue, and every annotation is carefully reviewed by at least three other experts before acceptance. In this way, we ensure the accuracy and rigor of the \textbf{Q-Bench$^+$} labels to the greatest extent possible. This, in turn, makes the performance testing capability of \textbf{Q-Bench$^+$} more precise and meaningful.

\subsection{\textbf{Distortions} and \textbf{Other Low-level Attributes} Enumeration}
\label{sec:a11}
\subsubsection{\textbf{Distortions}} Blurs [lens blur (out-of-focus), motion blur, zoom blur, gaussian blur, glass blur], Noises [gaussian noise, speckle noise, pepper noise], Artifacts [compression artifact, transmission error], Exposure Issues [under-exposure, over-exposure], Miscellaneous Artificial Distortions [pixelate, color-diffusion, jitter, \textit{etc}]
\subsubsection{\textbf{Other low-level attributes}} Color [color style, color vividity], Lighting [bright, dim], Composition [Symmetrical, Rule-of-Thirds], Visual Styles [animation, realism, computer-generated, AI-generated], Photographic Methods [background bokeh (shallow DOF), high contrast, motion blur (\textit{on fast-moving objects}), \textit{etc}]

\subsection{Multi-choice Question for Kosmos-2}
While Kosmos-2 performs generally well on the \textbf{description} and \textbf{assessment} tasks, we notice that it is hardly capable of answering a multi-choice question with the general prompt form applicable for other methods, as follows:
\\

\indent \indent \textit{{\small How is the clarity of the image? \\ \indent \indent  {\tt(Question)} [IMAGE\_TOKEN] {\tt(Image)} \\ \indent \indent  Choose between one of the following options: \\ \indent \indent  A. High {\tt{(Correct)}}  B. Medium{\tt(Wrong)}  C. Low{\tt(Wrong)}} } \\

 For most situations (\textbf{86\%}) in our primary sample test with the prompts above, Kosmos-2 will directly \textbf{append a new candidate} (\textit{e.g.,~\underline{D. Excellent} or \underline{D. Very Low}}) answer instead of choosing one option among them, denoted as \textbf{prompt failure}. This might be because the language model of Kosmos-2 has smaller capacity (1B) than other MLLMs that are based on LLaMA/MPT (7B/13B).

Considering that the prompt failure is actually not directly related with low-level perception, we try different prompt engineering techniques to reduce the prompt failure rate, and finalize with a simple modification which can limit the prompt failure to less than \textbf{10\%} in our sample set, as follows:
\\

\indent \indent \textit{{\small How is the clarity of the image? \\ \indent \indent  {\tt(Question)} [IMAGE\_TOKEN] {\tt(Image)} \\ \indent \indent  Choose between one of the following options:  \\ \indent \indent  A. High {\tt{(Correct)}}  B. Medium{\tt(Wrong)}  C. Low{\tt(Wrong)}}\\ \indent \indent  {\#Answer:}} \\

Nevertheless, we are still not able to eliminate the prompt failures for Kosmos-2. Henceforth, to systematically remove the negative effect of prompt failures on multi-choice questions for Kosmos-2, we conduct a choice-free special setting for it, \textit{i.e.} \textit{close-set} inference, via ranking the \textbf{perplexity} of different answers and choose the answer with minimum generative loss:
\\

\indent \indent \textit{{\small How is the clarity of the image? \\ \indent \indent  [IMAGE\_TOKEN] {\#Answer: High}}} \\ \indent \indent  $\to$ {\tt loss:7.43} $\to$ {\cmark~Choose this.} \\ \indent \indent 
\textit{{\small How is the clarity of the image? \\ \indent \indent [IMAGE\_TOKEN] {\#Answer: Medium}}} \\ \indent \indent  $\to$ {\tt loss:7.56} $\to$ \xmark  \\ \indent \indent  
\textit{{\small How is the clarity of the image? \\ \indent \indent  [IMAGE\_TOKEN] {\#Answer: Low}}} \\ \indent \indent  $\to$ {\tt loss:7.92} $\to$ \xmark \\

As shown in Table~\ref{tab:perplexity}, perplexity-based close-set inference can notably improve results of Kosmos-2. Considering that it is still the MLLM with fewest parameters among the ten models, its results are decent at its model size. More importantly, they validate that our observation of the prompt failure is reasonable.

\begin{table*}\small
    \centering
    \renewcommand\arraystretch{1.1}
    \renewcommand\tabcolsep{10pt}
    \caption{Perplexity-based \textit{close-set} evaluation compared with normal evaluation on the single images set of \textbf{LLVisionQA$^+$}; after eliminating the \textbf{prompt failures}, the results of Kosmos-2 significantly improved.}
    \vspace{-8pt}
    \resizebox{\linewidth}{!}{\begin{tabular}{l|ccc|cc|cc|c:c}
    \toprule
        \textbf{Sub-categories} & \multicolumn{3}{c|}{\textbf{Question Types}} & \multicolumn{4}{c|}{\textbf{Quadrants of Low-level Concerns}} & \multirow{3}{*}{{\textit{Overall$\uparrow$}}} & \multirow{3}{*}{{\textit{\#$\downarrow$}}}\\ \cdashline{1-8}
        \multirow{2}{*}{\textbf{Model} \textit{(variant)}}  & \multirow{2}{*}{\textit{Yes-or-No$\uparrow$}}& \multirow{2}{*}{\textit{What$\uparrow$}} & \multirow{2}{*}{\textit{How$\uparrow$}} & \multirow{2}{*}{\textit{Distortion$\uparrow$}} & \multirow{2}{*}{\textit{Other$\uparrow$}} & \textit{In-context}  &\textit{In-context} & &  \\
        &&&&&&\textit{Distortion$\uparrow$}& \textit{Other$\uparrow$} & & \\ \hline
        \textit{random guess} & 50.00\% & 28.18\% & 33.30\% & 37.54\% & 38.49\% & 38.70\% & 36.50\% & 37.87\% & -\\ \hdashline
        {$^{\star\star}$Kosmos-2 (normal)}  & 58.20\% & 29.13\% & 34.22\% & 38.10\% & 44.30\% & 40.93\% & 44.20\% & 41.47\% & \xmark \\
        {$^{\star\star}$Kosmos-2 (\textit{close-set})}  & \textbf{61.48\%} & \textbf{37.13\%} & \textbf{40.76\%} & \textbf{40.04\%} & \textbf{50.88\%} & \textbf{45.30\%} & \textbf{58.15\%} & \textbf{47.26\%} & \cmark \\
        \bottomrule
    \end{tabular}}
    \vspace{-5pt}
    \label{tab:perplexity}
\end{table*}

\begin{table*}\small
    \centering
    \renewcommand\arraystretch{1.1}
    \renewcommand\tabcolsep{10pt}
    \caption{Judgment accuracies of MLLMs on questions with correct answers as \textbf{Yes} or \textbf{No}.}
    \vspace{-8pt}
    \resizebox{0.8\linewidth}{!}{
    \begin{tabular}{l|c:cc:c:c}
    \toprule
        {\textbf{Model} \textit{(variant)}}  & \textit{all} & \textit{correct answer:}~\textbf{Yes} & \textit{correct answer:}~\textbf{No} & \textit{mean} & \textit{de-biased \#$\downarrow$} \\ 
        \hline
        \textit{random guess} & 50.00\% & 50.00\% & 50.00\% & 50.00\% & - \\ 
        \hdashline
        Shikra (\textit{Vicuna-7B}) & 66.91\% & 71.79\% & 60.00\% &  \textbf{65.90\%} & \textbf{3} \\
        LLaVA-v1 (\textit{Vicuna-13B}) & 57.10\% & 60.29\% & 51.66\% & 55.97\% & 6 \\
        MiniGPT-4 \textit{(Vicuna-13B)}  & 57.56\% & 70.00\% & 37.38\% &  53.69\% & 8 \\
        LLaMA-Adapter-V2  & 67.12\% & 68.80\% & 64.76\% & \textbf{66.78\%} & \textbf{2} \\
        InstructBLIP \textit{(Flan-T5-XL)}  & \textbf{68.67\%} & 80.14\% & 50.23\% & 65.19\%  & 4\\
        InstructBLIP \textit{(Vicuna-7B)}  & \textbf{71.40\%} & 84.32\% & 50.47\% & \textbf{67.39\%} & \textbf{1}\\
        Otter-v1 \textit{(MPT-7B)}  & 57.74\% & 70.14\% & 37.38\% & 53.76\% & 7 \\
        IDEFICS-Instruct \textit{(LLaMA-7B)}  & 59.74\% & 88.65\% & 13.09\% & 50.87\% & 9 \\
        mPLUG-Owl \textit{(LLaMA-7B)}  & \textbf{69.31\%} & 95.82\% & 26.67\% & 61.25\% & 5\\
    \bottomrule
    \end{tabular}
    }
    \vspace{-5pt}
    \label{tab:yesorno}
\end{table*}

\subsection{\textit{``Yes or No?''}: How Biased are MLLMs?}

In this section, we take a deeper analysis on the \textit{Yes-or-No} judgment ability of MLLMs, that whether these models can get similar accuracy on questions that should be answered with \textbf{Yes}, as those should be replied as \textbf{No}. Sadly, we notice that all MLLMs have higher prediction accuracy on \textbf{Yes}-questions than \textbf{No}-questions, while some MLLMs are more very severe biased (\textit{e.g.}, IDEFICS-Instruct). Considering that our \textbf{LLVisionQA} dataset contains more (62\%) \textbf{Yes}-questions than \textbf{No}-questions (38\%) and may introduce biases while comparing different MLLMs, we further compute a de-biased accuracy for all these methods, as the \textit{mean} value of the accuracies on two types of questions, and present the respective de-biased rank for all participating MLLMs, as listed in Table~\ref{tab:yesorno}. We hope this study on the biases and the de-biased results can provide a fairer comparison among them, as well as bring insights on the future improvements of MLLMs for low-level perception.

\subsection{Settings for GPT Evaluation of \textbf{Perception}} 

Given GPT's inherent variability, identical prompts can yield non-definitive responses. To address the impact of such situations on our evaluation, we've implemented a 5-round \textbf{voting} strategy. Under this approach, we pose the same prompt as defined in the following templates five times, taking the popular votes of GPT's answers to determine the final outcome. Our human analysis on a sample set confirms that the 5-round voting strategy improves GPT evaluation accuracy from \textbf{93.2\%} to \textbf{98.4\%}, reducing errors to only 1/4 compared with the single-round evaluation.

\textbf{Prompt Templates for GPT Evaluation:}

\textit{\small  \#System:~You are a helpful assistant that grades answers related to image quality and aesthetics. There are a lot of special terms or keywords related to image processing and photography. You will pay attention to the context of 'quality evaluation' when grading.} 

\textit{\small  \#User:~Assuming you are a grader, you will now be provided with a question [{question}] and a set of options [{options}] with option [{options[0]}] being the correct answer. Additionally, there will be an answer [{answer}] provided by a respondent. Please determine whether the respondent\'s answer is correct considering the context of the question. Even if the word choice is not completely the same, you can decide based on the given options and see whether the one in the answer is close enough to the given correct answer, The result is 1 if the answer is correct and else the result is 0. Please only provide the result in the following format: Result:} 

\subsection{Comparing natural images and AI-generated images}
{ There might be doubt that pairwise comparison between natural images (NIs) and AI-generated images (AIGIs) is unnecessary since the criteria for assessing quality differ between these two types of images is different. However, we believe it is essential to continue conducting comparisons between NIs and AIGIs for several reasons:}
\begin{itemize}
    \item {Common Technical Quality Issues: AIGIs are generated based on prior knowledge learned from NIs and, as such, they may inherit common technical issues such as noise, blur, and incorrect exposure. Therefore, examining and comparing the technical quality of NIs and AIGIs is meaningful, as it can guide and improve the technical quality of AIGIs.}
    \item {Aesthetic Quality Comparability: AIGIs and NIs also share comparability in aesthetic quality, such as in composition, color adjustment, and style. Investigating the aesthetic quality comparison between AIGIs and NIs can enhance the corresponding capabilities of MLLMs, thus improving the aesthetic quality of AI-generated content. Furthermore, considering the greater aesthetic freedom and diversity of AIGIs, comparing them with NIs can also provide more comprehensive aesthetic guidance for the photography of NIs.}
\end{itemize}

\subsection{Settings for GPT Evaluation of \textbf{Description}} 

The detailed \textbf{input prompt template} for GPT to evaluate the three dimensions is listed as follows:

\textit{\small  \#System:~You are a helpful assistant.} 
\begin{itemize}
    \item { \textbf{Completeness.} \textit{\small\#User: Evaluate whether the description [MLLM\_DESC] completely includes the low-level visual information in the reference description [GOLDEN\_DESC]. \\Please rate score 2 for completely or almost completely including reference information, 0 for not including at all, 1 for including part of the information or similar description. \\Please only provide the result in the following format: Score:}} 
    \item { \textbf{Preciseness.} \textit{\small\#User
    Evaluate whether the description [MLLM\_DESC] precisely reflects the reference description [GOLDEN\_DESC].  \\
    Please rate score 2 for totally no controversial low-level description, 1 for less controversial low-level description than matched descrpition, and 0 for more controversial low-level description than matched description. 
    Please only provide the result in the following format: Score: }}
    \item { \textbf{Relevance.} \textit{\small\#User: Evaluate whether the description [MLLM\_DESC] is relevant to the low-level visual information, which may include blur, noise, exposure, artifact, color, lighting, focus, composition, etc. \\Please rate score 2 for completely relevant, 1 for partly relevant, and 0 for totally irrelevant. \\Please only provide the result in the following format: Score:}}
\end{itemize}
In the prompt template, the \textit{[MLLM\_DESC]} denotes the output description from MLLMs, and \textit{[GOLDEN\_DESC]} denotes the \textit{golden} description in the \textbf{LLDescribe$^+$} dataset.

\begin{algorithm*}\footnotesize
\caption{{ Pytorch-style Pseudo Code for Softmax-based Strategy for IQA with MLLMs}}
\begin{lstlisting}
from PIL import Image
from my_mllm_model import Model, Tokenizer, embed_image_and_text

model, tokenizer = Model(), Tokenizer()

prompt = "##User: Rate the quality of the image.\n" \
         "##The quality of the image is \n Assistant: "

good_idx, poor_idx = tokenizer(["good","poor"]).tolist()

image = Image.open("image_for_iqa.jpg")
input_embeds = embed_image_and_text(image, prompt)
output_logits = model(input_embeds=input_embeds).logits[0,-1] 
q_pred = (output_logits[[good_idx, poor_idx]] / 100).softmax(0)[0]
\end{lstlisting}
\label{alg:1}
\end{algorithm*}

\begin{table*}[!htbp]\small
    \centering
    \renewcommand\arraystretch{1.5}
    \renewcommand\tabcolsep{6pt}
    \caption{{ Evaluation results on the \textit{prompt ensemble} strategy for the \textbf{Assessment} ability on MLLMs with top-7 results in the default A3 leaderboard of the \textbf{Q-Bench$^+$}. \textit{Prompt1} refers to \textit{satisfactory}+\textit{excellent}+\textit{perfect}$\leftrightarrow$\textit{terrible}+\textit{unsatisfactory}+\textit{subpar}. \textit{Prompt2} refers to \textit{good}+\textit{high}+\textit{fine}$\leftrightarrow$\textit{poor}+\textit{low}+\textit{bad}. Metrics are \textit{SRCC/PLCC}. Better performance in \textbf{BOLD}.} }
    \vspace{-5pt}
    \resizebox{\linewidth}{!}{\begin{tabular}{l|c|cccc|cc|c|c}
    \toprule
    {\textbf{Dataset Type}}  & \multirow{2}{*}{\textbf{Prompt}} & \multicolumn{4}{c|}{{In-the-wild}} & \multicolumn{2}{c|}{{Generated}} & \multicolumn{1}{c|}{{Artificial}} & \multirow{2}{27pt}{\textit{Average}}\\ \cdashline{1-1} \cdashline{3-9}
     \textbf{Model / Prompt / Dataset} & &{\textit{KONiQ-10k}} & {\textit{SPAQ}} & {\textit{LIVE-FB}} & \textit{LIVE-itw} & {\textit{CGIQA-6K}} & {\textit{AGIQA-3K}} & {\textit{KADID-10K}} & \\ \hline 
     \multirow{2}{*}{\textbf{InfiMM (\textit{Zephyr-7B})}~\cite{InfiMM}} & 
     \textit{Prompt1} & 0.471/0.541 & 0.633/0.610 & 0.257/0.276 & 0.576/0.576 & 0.263/0.261 & 0.731/0.775 & 0.491/0.461 & \textbf{0.489}/0.500 \\
     & \textit{Prompt2} & 0.496/0.533 & 0.605/0.618 & 0.266/0.291 & 0.569/0.593 & 0.239/0.248 & 0.708/0.768 & 0.492/0.490 & {0.482}/\textbf{0.506} \\  \hdashline
    \multirow{2}{*}{\textbf{Emu2-Chat (\textit{LLaMA-33B})}~\cite{emu2}} & 
     \textit{Prompt1} & 0.677/0.731 & 0.704/0.732 & 0.353/0.386 & 0.618/0.585 & 0.317/0.313 & 0.790/0.775 & 0.851/0.838 & 0.616/0.622\\
    &\textit{Prompt2} & {0.694}/{0.712} & {0.732}/{0.738} & {0.363}/{0.366} & {0.644}/0.613 & {0.321}/0.342 & 0.779/0.772 & 0.844/0.820& \textbf{0.625}/\textbf{0.623} \\  \hdashline
    
    \multirow{2}{*}{\textbf{InternLM-XComposer-VL (\textit{InternLM})}~\cite{xcomposer}} & 
     \textit{Prompt1} & 0.590/0.604 & 0.738/0.768 & 0.347/0.428 & 0.635/0.700 & 0.213/0.237 & 0.729/0.754 & 0.498/0.516 & 0.536/0.572\\
    &\textit{Prompt2} & 0.571/0.621 & 0.728/0.748 & 0.360/0.410 & 0.629/0.683  & 0.236/0.261 & 0.734/0.773 & 0.521/0.538& \textbf{0.540}/\textbf{0.576} \\ \hdashline
    
    \multirow{2}{*}{\textbf{LLaVA-v1.5 (\textit{Vicuna-v1.5-7B})}~\cite{improvedllava}} & 
     \textit{Prompt1} & 0.488/0.513 & 0.433/0.436 & 0.287/0.331 & 0.418/0.391 & 0.332/0.341 & 0.680/0.723 & 0.394/0.431 & 0.433/0.452 \\
    &\textit{Prompt2} & 0.512/0.513 & 0.443/0.465 & 0.303/0.315 & 0.408/0.415  & 0.318/0.324 & 0.697/0.752 & 0.392/0.421& \textbf{0.439}/\textbf{0.458} \\ \hdashline
    \multirow{2}{*}{\textbf{LLaVA-v1.5 (\textit{Vicuna-v1.5-13B})}~\cite{improvedllava}}  & \textit{Prompt1} & 0.464/0.524 & 0.545/0.608 & 0.288/0.358 & 0.515/0.512 & 0.311/0.347 & 0.688/0.762 & 0.412/0.415 & 0.460/\textbf{0.504} \\
     &\textit{Prompt2} & 0.474/0.498 & 0.565/0.588 & 0.314/0.345 & 0.488/0.521 & 0.311/0.322 & 0.692/0.771 & 0.382/0.392& \textbf{0.461}/0.491 \\ \hdashline

    \multirow{2}{*}{\textbf{Qwen-VL (\textit{QwenLM})}~\cite{Qwen-VL}} & 
     \textit{Prompt1} & 0.495/0.478 & 0.578/0.574 & 0.333/0.342 & 0.499/0.544 & 0.301/0.318 & 0.710/0.787 & 0.367/0.409 & 0.469/0.493\\
    &\textit{Prompt2} & 0.541/0.600 & 0.632/0.617 & 0.286/0.316 & 0.570/0.577 & 0.301/0.318 & 0.664/0.719 & 0.416/0.429& \textbf{0.487}/\textbf{0.511} \\ \hdashline 
    \multirow{2}{*}{\textbf{mPLUG-Owl (\textit{LLaMA-7B})}~\cite{mplugowl}} & 
     \textit{Prompt1} & 0.423/0.437 & 0.621/0.635 & 0.237/0.261 & 0.432/0.468 & 0.153/0.181 & 0.715/0.695 & 0.442/0.467 & \textbf{0.432}/0.449 \\
    &\textit{Prompt2} & 0.395/0.421 & 0.633/0.647 & 0.233/0.258 & 0.455/0.496  & 0.147/0.173 & 0.685/0.704& 0.463/0.483& 0.430/\textbf{0.455} \\ 
    
     \bottomrule
    \end{tabular}}
    \vspace{-10pt}
    \label{tab:complex}
\end{table*}

\begin{table*}[!t]\small
    \centering
    \renewcommand\arraystretch{1.3}
    \renewcommand\tabcolsep{7pt}
    \caption{{ Effectiveness of the proposed {\tt softmax} probability-based strategy against the baseline {\tt argmax} strategy, on multiple MLLMs and different IQA datasets. Metrics are \textit{SRCC/PLCC}. Higher in \textbf{BOLD}.}}
    \vspace{-5pt}
    \resizebox{\linewidth}{!}{\begin{tabular}{l|c|cccc|cc|c}
    \toprule
    {\textbf{Dataset Type}}  & & \multicolumn{4}{c|}{{In-the-wild}} & \multicolumn{2}{c|}{{Generated}} & \multicolumn{1}{c}{{Artificial}}\\ \hdashline
     \textbf{Model / Dataset} & Strategy  &{\textit{KONiQ-10k}} & {\textit{SPAQ}} & {\textit{LIVE-FB}} & \textit{LIVE-itw}  & \textit{CGIQA-6K} & {\textit{AGIQA-3K}} & {\textit{KADID-10K}}\\ \hline 
     Shikra (\textit{Vicuna-7B})~\cite{shikra}  & {\tt argmax}  & 0.178/0.201 & 0.277/0.281 & 0.152/0.169 & 0.248/0.267 & 0.071/0.065 & 0.513/0.562 & 0.245/0.246\\
     Shikra (\textit{Vicuna-7B})~\cite{shikra} & {\tt softmax} & \textbf{0.314/0.307} & \textbf{0.327/0.337} & \textbf{0.237/0.241} & \textbf{0.322/0.336}  & \textbf{0.198/0.201} & \textbf{0.640/0.661} & \textbf{0.324/0.332}\\
      \hdashline
        InstructBLIP \textit{(Vicuna-7B)}~\cite{iblip}  & {\tt argmax} & 0.284/0.352 & 0.662/0.664 & 0.156/0.249 & 0.195/0.264 &0.141/0.142 & 0.505/0.567 & 0.305/0.307 \\
        InstructBLIP \textit{(Vicuna-7B)}~\cite{iblip}  & {\tt softmax} & \textbf{0.359/\textbf{0.437}} & \textbf{0.683/\textbf{0.689}}  & \textbf{0.200/0.283} & \textbf{0.253/0.367}  & \textbf{0.263/0.304} & \textbf{0.629/0.663} &  \textbf{0.337/0.382}\\       \hdashline
        mPLUG-Owl \textit{(LLaMA-7B)}~\cite{mplugowl}  & {\tt argmax} & 0.111/0.154 & 0.463/0.469 & 0.081/0.123 & 0.169/0.237 &0.082/0.067 & 0.410/0.466 & 0.203/0.204  \\
        mPLUG-Owl \textit{(LLaMA-7B)}~\cite{mplugowl}  & {\tt softmax} & \textbf{0.409/0.427} & \textbf{0.634/\textbf{0.644}} & \textbf{0.241/0.271} & \textbf{0.437/0.487} & \textbf{0.148/0.180} & \textbf{0.687/0.711} &  \textbf{0.466/0.486} \\   \hdashline
        LLaMA-Adapter-V2~\cite{llamaadapterv2} & {\tt argmax} & 0.218/0.237 &  0.417/0.423 & 0.222/0.257 & 0.205/0.239 & 0.152/0.116 & 0.545/0.579 & 0.228/0.229 \\
        LLaMA-Adapter-V2~\cite{llamaadapterv2} & {\tt softmax} & \textbf{0.354/0.363} & \textbf{0.464/0.506} &  \textbf{0.275/0.329} & \textbf{0.298/0.360} &\textbf{0.251/0.257} & \textbf{0.604/0.666} &  \textbf{0.412/0.425} \\       \hdashline
        LLaVA-v1 \textit{(Vicuna-13B)}~\cite{llava}  & {\tt argmax} & 0.038/0.045 & 0.101/0.108 & 0.036/0.035 & 0.059/0.075 &0.112/0.109 & 0.240/0.297 & 0.005/0.005 \\
        LLaVA-v1 \textit{(Vicuna-13B)}~\cite{llava} & {\tt softmax}& \textbf{0.462/\textbf{0.457}} & \textbf{0.442/0.462} & \textbf{0.264/0.280} & \textbf{0.404/0.417}  & \textbf{0.285/0.297} & \textbf{0.626/\textbf{0.684}} & \textbf{0.349/0.372} \\      
     \bottomrule
    \end{tabular}}
    \vspace{-10pt}
    \label{tab:assessment_deepdive}
\end{table*}

\begin{table*}[!t]\small
    \centering
    \renewcommand\arraystretch{1}
    \renewcommand\tabcolsep{6pt}
    \caption{{ Evaluation results on the \textit{prompt ensemble} strategy for the \textbf{Assessment} ability on MLLMs with top-7 results in the default A3 leaderboard of the \textbf{Q-Bench$^+$}. After \textit{ensemble}, the rankings among them are not changed. Metrics are \textit{SRCC/PLCC}. Best in \textbf{BOLD} and second \underline{uderlined}.}}
    \vspace{-5pt}
    \resizebox{\linewidth}{!}{\begin{tabular}{l|cccc|cc|c|c}
    \toprule
    {\textbf{Dataset Type}}  & \multicolumn{4}{c|}{{In-the-wild}} & \multicolumn{2}{c|}{{Generated}} & \multicolumn{1}{c|}{{Artificial}} & \multirow{2}{27pt}{\textit{Average}}\\ \cdashline{1-8}
     \textbf{Prompt / Dataset}  &{\textit{KONiQ-10k}} & {\textit{SPAQ}} & {\textit{LIVE-FB}} & \textit{LIVE-itw} & {\textit{CGIQA-6K}} & {\textit{AGIQA-3K}} & {\textit{KADID-10K}} & \\ \hline 
         \textbf{InfiMM (\textit{Zephyr-7B})}~\cite{InfiMM} \\ \hdashline
    \textit{good}$\leftrightarrow$\textit{poor}  & 0.507/0.546 & 0.616/0.633 & 0.268/0.299 & 0.548/0.580 & 0.229/0.245 & 0.706/0.767 & 0.466/0.452 & \underline{0.477}/\underline{0.503} \\
    \textit{fine}$\leftrightarrow$\textit{bad} & 0.331/0.368 & 0.500/0.527 & 0.190/0.251 & 0.305/0.366 & 0.309/0.324 & 0.555/0.651 & 0.411/0.430 & 0.372/0.417 \\
    \textit{high}$\leftrightarrow$\textit{low}  & 0.412/0.382 & 0.539/0.492 & 0.216/0.194 & 0.586/0.524 & 0.173/0.171 & 0.674/0.698 & 0.429/0.429 & 0.433/0.413 \\
    \textit{good}+\textit{high}$\leftrightarrow$\textit{poor}+\textit{low}  & 0.475/0.492 & 0.589/0.583 & 0.249/0.253 & 0.582/0.578 & 0.198/0.201 & 0.697/0.750 & 0.454/0.456 & 0.463/0.473 \\
    \textit{good}+\textit{fine}$\leftrightarrow${poor}+{bad}  & 0.463/0.502 & 0.591/0.613 & 0.255/0.299 & 0.488/0.530 & 0.272/0.287 & 0.675/0.749 & 0.479/0.467 & 0.460/0.493 \\
    \textit{good}+\textit{high}+\textit{fine}$\leftrightarrow$\textit{poor}+\textit{low}+\textit{bad} & 0.496/0.533 & 0.605/0.618 & 0.266/0.291 & 0.569/0.593 & 0.239/0.248 & 0.708/0.768 & 0.492/0.490 & \textbf{0.482}/\textbf{0.506} \\ \hline

    \textbf{Emu2-Chat (\textit{LLaMA-33B})}~\cite{emu2} \\ \hdashline
    \textit{good}$\leftrightarrow$\textit{poor} & {0.664}/{0.714} & {0.712}/{0.698} & {0.355}/{0.341} & {0.597}/{0.611} & 0.224/0.269 & {0.759}/{0.751} & {0.841}/{0.790} & {0.593}/{0.596} \\
    \textit{fine}$\leftrightarrow$\textit{bad} & 0.663/0.540 & 0.711/0.702 & 0.359/{0.362} & 0.601/0.631 & 0.285/0.334 & 0.770/0.599 & 0.846/0.830& 0.605/0.571\\
    \textit{high}$\leftrightarrow$\textit{low} & 0.685/0.644 & 0.721/0.703 & 0.333/0.334 & 0.633/0.647 & 0.255/0.237 & 0.779/0.793 & 0.830/0.795& 0.605/0.593\\
    \textit{good}+\textit{high}$\leftrightarrow$\textit{poor}+\textit{low} & {0.696/0.732} & {0.744}/0.721 & 0.341/0.320 & {0.656}/{0.671} & {0.307}/{0.347} & 0.775/0.796 & 0.841/0.794& \underline{0.622}/\textbf{0.625}\\
    \textit{good}+\textit{fine}$\leftrightarrow${poor}+{bad} & 0.674/0.678 & 0.731/{0.735} & {0.360}/0.356 & 0.632/{0.654} & 0.298/0.343 & 0.771/0.743 & 0.847/0.830& 0.616/0.619 \\
    \textit{good}+\textit{high}+\textit{fine}$\leftrightarrow$\textit{poor}+\textit{low}+\textit{bad} & {0.694}/{0.712} & {0.732}/{0.738} & {0.363}/{0.366} & {0.644}/0.613 & {0.321}/0.342 & 0.779/0.772 & 0.844/0.820& \textbf{0.625}/\underline{0.623} \\  \hline

    \textbf{InternLM-XComposer-VL (\textit{InternLM})}~\cite{xcomposer} \\ \hdashline
    \textit{good}$\leftrightarrow$\textit{poor} & 0.564/0.615 & 0.730/0.750 & 0.360/0.416 & 0.612/0.676  & 0.243/0.265 & 0.732/0.775 & 0.546/0.572& \underline{0.541}/\textbf{0.581}\\
    \textit{fine}$\leftrightarrow$\textit{bad} & 0.546/0.597 & 0.720/0.736 & 0.341/0.389 & 0.626/0.671  & 0.213/0.227 & 0.681/0.708 & 0.494/0.479& 0.517/0.544\\
    \textit{high}$\leftrightarrow$\textit{low} & 0.543/0.590 & 0.704/0.720 & 0.331/0.372 & 0.612/0.656  & 0.223/0.251 & 0.716/0.755 & 0.490/0.500& 0.517/0.549\\
    \textit{good}+\textit{high}$\leftrightarrow$\textit{poor}+\textit{low} & 0.564/0.613 & 0.723/0.743 & 0.354/0.405 & 0.621/0.676  & 0.238/0.264 & 0.734/0.775 & 0.522/0.546& 0.537/0.575\\
    \textit{good}+\textit{fine}$\leftrightarrow$\textit{poor}+\textit{bad} & 0.573/0.626 & 0.735/0.755 & 0.366/0.420 & 0.629/0.687  & 0.236/0.260 & 0.732/0.771 & 0.531/0.551& \textbf{0.543/0.581} \\
    \textit{good}+\textit{high}+\textit{fine}$\leftrightarrow$\textit{poor}+\textit{low}+\textit{bad} & 0.571/0.621 & 0.728/0.748 & 0.360/0.410 & 0.629/0.683  & 0.236/0.261 & 0.734/0.773 & 0.521/0.538& 0.540/0.576 \\ \hline
    
    \textbf{LLaVA-v1.5 (\textit{Vicuna-v1.5-7B})}~\cite{improvedllava} \\ \hdashline
    \textit{good}$\leftrightarrow$\textit{poor} & 0.463/0.459 & 0.443/0.467 & 0.305/0.321 & 0.344/0.358  & 0.321/0.333 & 0.672/0.738 & 0.417/0.440& {0.424}/{0.445} \\
    \textit{fine}$\leftrightarrow$\textit{bad} & 0.453/0.469 & 0.457/0.482 & 0.258/0.288 & 0.303/0.333  & 0.294/0.302 & 0.558/0.617 & 0.389/0.420& 0.388/0.416 \\
   \textit{high}$\leftrightarrow$\textit{low} & 0.474/0.476 & 0.370/0.386 & 0.261/0.262 & 0.432/0.429  & 0.266/0.269 & 0.669/0.716 & 0.304/0.331& 0.397/0.410 \\
    \textit{good}+\textit{high}$\leftrightarrow$\textit{poor}+\textit{low} & 0.491/0.491 & 0.416/0.436 & 0.293/0.300 & 0.696/0.751 & 0.413/0.416  & 0.298/0.304 & 0.359/0.389& 0.424/0.441 \\
    \textit{good}+\textit{fine}$\leftrightarrow$\textit{poor}+\textit{bad} & 0.482/0.482 & 0.461/0.485 & 0.300/0.320 & 0.644/0.708 & 0.339/0.357  & 0.327/0.336 & 0.425/0.451& \underline{0.425}/\underline{0.449} \\
    \textit{good}+\textit{high}+\textit{fine}$\leftrightarrow$\textit{poor}+\textit{low}+\textit{bad} & 0.512/0.513 & 0.443/0.465 & 0.303/0.315 & 0.408/0.415  & 0.318/0.324 & 0.697/0.752 & 0.392/0.421& \textbf{0.439/0.458} \\ \hline
    \textbf{LLaVA-v1.5 (\textit{Vicuna-v1.5-13B})}~\cite{improvedllava} \\ \hdashline
     \textit{good}$\leftrightarrow$\textit{poor} & 0.448/0.460 & 0.563/0.584 & 0.310/0.339 & 0.445/0.481  & 0.285/0.297 & 0.664/0.754 & 0.390/0.400& 0.444/0.473 \\
     \textit{fine}$\leftrightarrow$\textit{bad} & 0.449/0.487 & 0.583/0.597 & 0.316/0.360 & 0.466/0.513  & 0.349/0.365 & 0.650/0.749 & 0.425/0.437& \underline{0.463}/\textbf{0.501} \\
    \textit{high}$\leftrightarrow$\textit{low} & 0.456/0.482 & 0.529/0.553 & 0.286/0.306 & 0.489/0.513  & 0.276/0.284 & 0.683/0.752 & 0.316/0.331& 0.434/0.460 \\
    \textit{good}+\textit{high}$\leftrightarrow$\textit{poor}+\textit{low} & 0.462/0.484 & 0.548/0.573 & 0.303/0.327 & 0.480/0.509  & 0.283/0.294 & 0.687/0.763 & 0.350/0.363& 0.445/0.473 \\
    \textit{good}+\textit{fine}$\leftrightarrow$\textit{poor}+\textit{bad} & 0.463/0.483 & 0.579/0.596
    & 0.321/0.356 & 0.467/0.505 & 0.326/0.339 & 0.670/0.762 & 0.420/0.426& \textbf{0.464}/\underline{0.495} \\
    \textit{good}+\textit{high}+\textit{fine}$\leftrightarrow$\textit{poor}+\textit{low}+\textit{bad} & 0.474/0.498 & 0.565/0.588 & 0.314/0.345 & 0.488/0.521 & 0.311/0.322 & 0.692/0.771 & 0.382/0.392& 0.461/0.491 \\ \hline
    \textbf{Qwen-VL (\textit{QwenLM})}~\cite{Qwen-VL}\\ \hdashline 
    \textit{good}$\leftrightarrow$\textit{poor} & 0.470/0.546 & 0.676/0.669 & 0.298/0.339 & 0.504/0.532  & 0.273/0.284 & 0.617/0.686 & 0.486/0.486& {0.475}/{0.506} \\
    \textit{fine}$\leftrightarrow$\textit{bad} & 0.467/0.507 & 0.352/0.365 & 0.205/0.238 & 0.451/0.472  & 0.188/0.185 & 0.599/0.627 & 0.354/0.378& 0.374/0.396 \\
    \textit{high}$\leftrightarrow$\textit{low} & 0.531/0.578 & 0.626/0.616 & 0.281/0.290 & 0.574/0.560  & 0.286/0.314 & 0.637/0.692 & 0.332/0.344& 0.467/0.485 \\
    \textit{good}+\textit{high}$\leftrightarrow$\textit{poor}+\textit{low} & 0.539/0.600 & 0.684/0.673 & 0.299/0.324 & 0.565/0.568  & 0.306/0.330 & 0.660/0.721 & 0.414/0.422& \textbf{0.495/0.520} \\
    \textit{good}+\textit{fine}$\leftrightarrow$\textit{poor}+\textit{bad} & 0.495/0.558 & 0.596/0.581 & 0.264/0.307 & 0.521/0.548  & 0.270/0.270 & 0.640/0.691 & 0.435/0.449& 0.460/0.486\\
    \textit{good}+\textit{high}+\textit{fine}$\leftrightarrow$\textit{poor}+\textit{low}+\textit{bad} & 0.541/0.600 & 0.632/0.617 & 0.286/0.316 & 0.570/0.577 & 0.301/0.318 & 0.664/0.719 & 0.416/0.429& \underline{0.487}/\underline{0.511} \\ \hline
    \textbf{mPLUG-Owl (\textit{LLaMA-7B})}~\cite{mplugowl} \\ \hdashline 
    \textit{good}$\leftrightarrow$\textit{poor} & 0.409/0.427 & 0.634/0.644 & 0.241/0.271 & 0.437/0.487  & 0.148/0.180 & 0.687/0.711 & 0.466/0.486& \underline{0.432}/\underline{0.458}\\
    \textit{fine}$\leftrightarrow$\textit{bad} & 0.357/0.398 & 0.622/0.636 & 0.260/0.290 & 0.422/0.475  & 0.178/0.224 & 0.606/0.646 & 0.536/0.534& 0.426/0.457\\
    \textit{high}$\leftrightarrow$\textit{low} & 0.353/0.369 & 0.610/0.624 & 0.176/0.187 & 0.436/0.464 & 0.110/0.124 & 0.662/0.663 & 0.361/0.378& 0.387/0.401\\
    \textit{good}+\textit{high}$\leftrightarrow$\textit{poor}+\textit{low} & 0.382/0.402 & 0.626/0.642 & 0.208/0.228 & 0.446/0.483  & 0.125/0.144 & 0.684/0.697 & 0.409/0.432& 0.411/0.432\\
    \textit{good}+\textit{fine}$\leftrightarrow$\textit{poor}+\textit{bad} & 0.403/0.430 & 0.635/0.645 & 0.260/0.292 & 0.444/0.493  & 0.172/0.213 & 0.664/0.694 & 0.525/0.527& \textbf{0.443/0.471} \\
    \textit{good}+\textit{high}+\textit{fine}$\leftrightarrow$\textit{poor}+\textit{low}+\textit{bad} & 0.395/0.421 & 0.633/0.647 & 0.233/0.258 & 0.455/0.496  & 0.147/0.173 & 0.685/0.704& 0.463/0.483& 0.430/0.455 \\ 
    
     \bottomrule
    \end{tabular}}
    \vspace{10pt}
    \label{tab:assessment_ensemble}
\end{table*}

\subsection{Evaluation Details for \textbf{Assessment} Ability}
\subsubsection{Example pseudo code for MLLMs on IQA}

{ In Algo.~\ref{alg:1}, we provide an example of how to evaluate image quality with MLLMs. The algorithm is simple with \textit{only 9 lines}, and could be easily integrated with any new MLLMs (\textit{based on causal LLMs}), so as to allow these models to quantitatively predict the quality of images.}

\subsubsection{Detailed performance for {\tt softmax} \& prompt ensemble strategy}
{ In this section, we present a detailed analysis of the performance of the {\tt softmax} and prompt ensemble strategies. This comprehensive evaluation aims to provide a precise comparison and demonstrate the effectiveness of the proposed strategies. The  {\tt softmax} and {\tt argmax} comparison is illustrated in Table~\ref{tab:assessment_deepdive} while the prompt ensemble performance is exhibited in Table~\ref{tab:assessment_ensemble}.}

\subsubsection{Enriching prompt set}
{ We concur that incorporating more complex synonyms in the prompt ensemble strategy can help mitigate word bias and enable the prompt-ensemble strategy to more accurately target \textit{positive} or \textit{negative} quality embeddings. At the same time, we remain open to exploring different choices for prompt ensembles. The selection of terms in our paper represents a preliminary attempt, thus our choices are intentionally simplistic, aimed only at demonstrating the effectiveness of the prompt-ensemble strategy. Further enriching this selection is a valuable direction in prompt engineering, potentially unlocking the full assessment potential of the model. Therefore, we follow your suggestions by enriching the prompts with more complex words: For the \textit{positive} prompts, we employ the set \{\textit{satisfactory, excellent, perfect}\}, and for the \textit{negative} prompts, the set \{\textit{terrible, unsatisfactory, subpar}\} is utilized. The outcomes are depicted in Table \ref{tab:complex}. From this table, we observe that using complex prompts yields performance results comparable to those achieved with simple prompts. This suggests that the contributions of complex and simple prompts within the prompt-ensemble strategy are not significantly different.}

\begin{table*}[!t]\small
    \centering
    \renewcommand\arraystretch{1.1}
    \renewcommand\tabcolsep{7.5pt}
        \caption{Performance of MLLMs that are fine-tuned with low-level multi-modal dataset Q-Instruct~\cite{q-instruct} on the {\tt test} subset of \textbf{Q-Bench$^+$}. \textit{Perception}, \textit{Description}, and \textit{Assessment} are the overall accuracy, the sum of the description scores, and average \textit{SRCC/PLCC} values respectively.}
        \vspace{-8pt}
    \resizebox{.7\linewidth}{!}{\begin{tabular}{l|cc|cc|c}
    \toprule
        \textbf{Dimensions} & \multicolumn{2}{c|}{\textbf{Single}} & \multicolumn{2}{c|}{\textbf{Pair}} & \multirow{2}{*}{\textit{Assessment}} \\ \cdashline{1-5}
        \textbf{Model} (\textit{variant}) & \textit{Perception} & \textit{Description} & \textit{Perception} & \textit{Description} & \\ \hline
        \multicolumn{6}{l}{\textit{Without Q-Instruct/Co-Instruct}} \\ \hdashline
        mPLUG-Owl2 (\textit{LLaMA-7B})~\cite{mplug2} & 62.68\% & 3.67 & 48.94\% & 3.50 & 0.326/0.357\\
        LLaVA-v1.5 (\textit{Vicuna-v1.5-7B})~\cite{improvedllava} & 60.07\% & 3.21  & 52.25\% & 3.22 & 0.424/0.445\\
        LLaVA-v1.5 (\textit{Vicuna-v1.5-13B})~\cite{improvedllava} & 61.40\% & 3.47 & 52.05\% & 3.44 & 0.444/0.474\\
        InternLM-XComposer-VL \textit{(InternLM)}~\cite{xcomposer} & 64.35\% & 4.21 & 51.11\% & 3.51 & 0.541/0.581\\
        \hline
        \multicolumn{5}{l}{\textit{With Q-Instruct}} \\ \hdashline
        mPLUG-Owl2 \textit{(LLaMA-7B)}~\cite{mplug2} & 69.10\% & 3.99 & 51.22\% & 3.69 & \textbf{0.727/0.742} \\
        LLaVA-v1.5 (\textit{Vicuna-v1.5-7B})~\cite{improvedllava} & 67.42\% & 3.82 & 53.17\% & 3.41 & 0.691/0.722\\
        LLaVA-v1.5 (\textit{Vicuna-v1.5-13B})~\cite{improvedllava} & 70.43\% & 4.00 & 54.44\% & 3.57 & 0.649/0.677\\
        InternLM-XComposer-VL \textit{(InternLM)}~\cite{xcomposer} & 70.37\% & 4.25 & 53.21\% & 3.66 & 0.686/0.711\\ \hline
        \multicolumn{5}{l}{\textit{With Co-Instruct}} \\ \hdashline
        mPLUG-Owl2 \textit{(LLaMA-7B)}~\cite{mplug2} & \textbf{77.12\%} & \textbf{4.26} & \textbf{80.18\%} & \textbf{4.82} & 0.716/0.739 \\ \hline
        Best \textit{Open-Source} Performance & 67.69\% & 4.21 & 53.15\% & 3.50 & 0.593/0.596\\
        Best \textit{Closed-Source} Performance & 74.10\% & - & 78.07\% & - & -/-\\
         \bottomrule
    \end{tabular}}
    \vspace{-10pt} 
    \label{tab:finetune}
\end{table*}

\subsubsection{IQA evaluation strategy for CLIP-ViT-Large-14}

In Tab.~\ref{tab:assessment}, we compare the IQA performance of MLLMs with CLIP-ViT-Large-14, the visual backbone of the majority of MLLMs. Attempting to understand whether the new language part (LLM) can do better than the original language part of CLIP, we try to compare between CLIP and MLLMs in a relatively \textbf{aligned} setting. Firstly, noticing that most MLLMs will resize images into $224\times224$ as their input sizes, we align this setting on CLIP, and ignore the strategies as proposed by~\cite{clipiqa}. Secondly, same as the strategy on MLLMs, we also apply {\tt softmax} pooling between \textbf{\textit{good}} and \textbf{\textit{poor}}, as in the CLIP's zero-shot classification format: \textit{a photo of good quality} and \textit{a photo of poor quality}. Besides the two alignments, the quality scores of CLIP-ViT-Large-14 are obtained as follows:
\begin{equation}
q_{clip} \! = \!  \frac{e^{\mathrm{CS}(f_{[\text{IMAGE}]}, f_{\textbf{a photo of good quality}})}}{e^{\mathrm{CS}(f_{[\text{IMAGE}]}, f_{\textbf{a photo of good quality}})}+e^{\mathrm{CS}(f_{[\text{IMAGE}]}, f_{\textbf{a photo of poor quality}})}}
\end{equation}
where $CS(\cdot)$ represents the cosine similarity calculation function.

\subsubsection{Special IQA settings for Flan-T5-based InstructBLIP}

For InstructBLIP~\cite{iblip} (\textit{Flan-T5-XL}), different from the majority of LLaMA-based (or MPT-based Otter-v1) MLLMs, the two top-frequency tokens are \textbf{\textit{high}} (89\%) and \textbf{\textit{low}} (8\%) instead of the common \textbf{\textit{good$\leftrightarrow$poor}}. Henceforth, based on our motivation to only modify the {\tt argmax} into {\tt softmax} and follow the default \textbf{top-frequency} output tokens of MLLMs, we replace the probabilities of \textbf{\textit{good$\leftrightarrow$poor}} into those of \textbf{\textit{high$\leftrightarrow$low}} in Eq.~\ref{eq:1} for T5, defined as follows:

\begin{equation}
q_\mathrm{pred,T5} = \frac{e^{x^\text{\textbf{high}}_{\textit{SCORE\_TOKEN}}}}{e^{x^\text{\textbf{high}}_{\textit{SCORE\_TOKEN}}}+e^{x^\text{\textbf{low}}_{\textit{SCORE\_TOKEN}}}}
\label{eq:t5}
\end{equation}

As validated in our experiments~(Table~\ref{tab:assessment_ensemble}, the \textbf{\textit{high$\leftrightarrow$low}} pair generally predicts better than \textbf{\textit{good$\leftrightarrow$poor}} on majority of databases. The better performance on \textbf{MLLM-specific top-frequency tokens} by side validates the effectiveness of our methodology for MLLMs on IQA.

\subsubsection{Special validation protocol for CGIQA-6K}

The CGIQA-6K ~\cite{zhang2023subjective} dataset contains two separate sub-sets which consist of 3,000 game images and 3,000 movie images respectively, with \textbf{different instructions} for human annotators during its subjective experiments. Therefore, we validate the MLLMs' assessment performance on the two sub-sets individually and average the results for the final exhibition. The results of NIQE and CLIP-ViT-Large-14 are also obtained under the same protocol for a fair comparison. 

\subsection{MLLMs Finetuned with Low-level Dataset}
{ Additionally, we have proceeded to conduct the corresponding experiments separately to compare the performance of MLLMs that are fine-tuned on low-level multi-modal datasets. Q-Instruct~\cite{q-instruct} is the first large-scale, low-level multi-modal dataset available (focusing on single images), prompting our decision to fine-tune four popular MLLMs (LLaVA-v1.5 (\textit{Vicuna-v1.5-7B}), LLaVA-v1.5 (\textit{Vicuna-v1.5-13B}), mPLUG-Owl2 \textit{(LLaMA-7B)}, and InternLM-XComposer-VL \textit{(InternLM)}.) using this dataset.  More recently, the Co-Instruct~\cite{co_instruct} dataset is proposed to cover low-level data under both single and comparison settings, which is utilized to fine-tune mPLUG-Owl2 \textit{(LLaMA-7B)}. The experimental results are presented in Table \ref{tab:finetune}. a) From the table, we observe that fine-tuning with Q-Instruct significantly enhances performance on single image tasks (all finetuned MLLMs are superior to the best \textit{open-source} performance), but only slightly improves performance on image pair tasks. This suggests that while the low-level multi-modal dataset covering only single images helps MLLMs acquire better low-level knowledge, it does not sufficiently enhance their comparison abilities, resulting in limited performance gains for image pair tasks. b) Conversely, the Co-Instruct dataset markedly improves performance for both single images and image pair tasks. This improvement indicates that incorporating training with low-level image pair data is advantageous for tasks involving image comparisons.}

\subsection{Statement on Data Contamination}
The \textbf{Q-bench$^+$} contains three tasks, where the first two tasks, (A1) \textbf{perception} and (A2) \textbf{description}, are evaluated with our own datasets proposed with the paper. For these two tasks, the questions, answers, or low-level descriptions in the two datasets are not seen by any existing MLLMs. Half of \textbf{LLVisionQA$^+$} (\textit{i.e.} the {\tt test} subset) and full of \textbf{LLDescribe$^+$} labels are kept private, to avoid being added to the training sets of any MLLMs. We hope that this measure will allow \textbf{Q-bench$^+$} to have long-term significance as an indicator of low-level visual abilities.

For the third task, (A3) \textbf{assessment}, the situation is a bit more complicated. 
For open-source models as tested, almost all of them have provided their technical reports, and as far as we know, \textbf{no} image quality assessment (IQA)  dataset has participated in the \textbf{multi-modality training stages} of them. While text knowledge about image quality assessment should have been injected to them (\textit{e.g.} a blurry image is a low quality image) during their \textbf{pure-language training stages}, we think this should not be regarded as data contamination for IQA, because the images cannot be seen by a language model. Instead, they are important knowledge for MLLMs to better link particular visual attributes (\textit{blur}) to human opinions (\textit{quality}), which motivates us to explore MLLMs for these tasks.

\end{document}